\documentclass[draft]{agujournal2019}
\usepackage{xr}
\usepackage{url} 
\usepackage{siunitx}
\newcommand{\NA}{---}
\draftfalse

\usepackage{multirow}
\usepackage{amssymb}
\usepackage{xcolor}
\usepackage{float}
\usepackage{comment}
\usepackage{gensymb}
\usepackage{textcomp}
\usepackage{amsmath}
\usepackage{graphicx}
\usepackage{array}
\usepackage{booktabs}
\usepackage{longtable}
\usepackage{threeparttable}
\usepackage[finalnew]{trackchanges} 
\usepackage{soul}

\usepackage{makecell}

\journalname{Water Resources Research}

\begin{document}


%


\title{Predicting Water Temperature Dynamics of Unmonitored Lakes with Meta Transfer Learning}
%
%




\authors{
Jared D. Willard\affil{1,2} , Jordan S. Read\affil{2}, Alison P. Appling\affil{2}, Samantha K. Oliver\affil{2}, Xiaowei Jia\affil{3}, and Vipin Kumar\affil{1}}

\affiliation{1}{Department of Computer Science and Engineering, University of Minnesota, Minneapolis, MN, USA}
\affiliation{2}{U.S. Geological Survey, Middleton, WI, USA
}
\affiliation{3}{Department of Computer Science, University of Pittsburgh, Pittsburgh, PA, USA}





\correspondingauthor{Jared D. Willard}{willa099@umn.edu}




\begin{keypoints}
\item Meta Transfer Learning (MTL) learns from models trained on data-rich systems to inform predictions in systems where no observations exist 
\item We use MTL with process-based and process-guided deep learning models to accurately predict lake temperatures in the Midwest United States
\item The most important predictor of transfer model success is the difference in maximum depth between the data-rich and \change{unobserved}{unmonitored} lake
\end{keypoints}
%

%
%

%
%

\begin{abstract}
Most environmental data come from a minority of well-\change{observed}{monitored} sites. An ongoing challenge in the environmental sciences is transferring knowledge from monitored sites to \change{unobserved}{unmonitored} sites. Here, we demonstrate a novel transfer learning framework that accurately predicts \add{depth\protect-specific} temperature in \change{unobserved}{unmonitored} lakes (targets) by borrowing models from \change{highly observed}{well-monitored} lakes (sources). This method, Meta Transfer Learning (MTL), builds a meta-learning model to predict transfer performance from candidate source models to targets using lake attributes and candidates’ past performance. We constructed source models at 145 \change{well-observed}{well-monitored} lakes using calibrated process-based modeling (PB) and a recently developed approach called process-guided deep learning (PGDL). We applied MTL to either PB or PGDL source models (PB-MTL or PGDL-MTL, respectively) to predict temperatures in 305 target lakes treated as \change{unobserved}{unmonitored} in the Upper Midwestern United States. We show significantly improved performance relative to the uncalibrated process-based General Lake Model, where the median RMSE for the target lakes is 2.52 \textdegree C. PB-MTL yielded a median RMSE of 2.43 \textdegree C; PGDL-MTL yielded 2.16 \textdegree C; and a PGDL-MTL ensemble of nine sources per target yielded 1.88 \textdegree C. For sparsely \change{observed}{monitored} target lakes, PGDL-MTL often outperformed PGDL models trained on the target lakes themselves. Differences in maximum depth between the source and target were consistently the most important predictors. Our approach readily scales to thousands of lakes in the Midwestern United States, demonstrating that MTL with meaningful predictor variables and high-quality source models is a promising approach for many kinds of unmonitored systems and environmental variables.

\end{abstract}





%
\section{Introduction}
Environmental data often does not exist at the appropriate resolution or extent for decision making or characterizing change. Models can be used to fill gaps in key ecosystem variables, such as extreme precipitation rates \cite{lockhoff2014evaluation}, soil moisture \cite{mishra2017drought}, hydrological flow \cite{liu2017review}, and lake temperature \cite{aguilera2016using}, which otherwise would be unavailable at the spatial and temporal scales needed for ecological decision-making \cite{lovett2007needs}. Although sensor data is increasingly prevalent, it will always be incomplete, especially for variables where observations are concentrated in a small subset of locations and the majority of locations remain unmonitored. Since observing key variables like these at scale is prohibitively costly \cite{caughlan2001cost}, models that can efficiently use existing data and transfer information to unmonitored systems are critical to closing our information gaps. 

\note{Added Table 1 of abbreviations}
\begin{table}[h]
\centering
\caption{\textit{List of Abbreviations}}
\scalebox{0.8}{
\begin{tabular}{ |l|l| } 
\hline
Abbreviation & Definition \\
\hline \hline
CV & Cross Validation \\
\hline
GLM & General Lake Model \\
\hline
LSTM & Long Short-term Memory\\
\hline
ML & Machine Learning \\
\hline
MSE & Mean Square Error \\
\hline
MTL & Meta Transfer Learning \\ 
\hline
PB & Calibrated Process-Based Model \\
\hline
PB0 & Uncalibrated Process-Based Model \\
\hline
PB-MTL & Process-Based Model Meta Transfer Learning \\
\hline
PGDL & Process-Guided Deep Learning \\ 
\hline
PGDL-MTL & \thead{Process-Guided Deep Learning Model Meta Transfer \\Learning} \\
\hline
PGDL-MTL9 & \thead{Process-Guided Deep Learning Model Meta Transfer \\Learning with Ensemble of 9 Source Models}\\
\hline
RFECV & Recursive Feature Elimination with Cross Validation\\
\hline
RMSE & Root Mean Square Error\\
\hline
$r_s$ & Spearman Rank Correlation Coefficient \\
\hline
\end{tabular}}
\label{tab:abbrev}
\end{table}


There are many modeling approaches for predicting complex environmental phenomena, and model choice can be viewed as a trade-off among prediction accuracy, data needs, and generalizability to new systems. Process-based models are a popular modeling choice for water resources tasks like the prediction of stream temperature \cite{dugdale2017river}, hydrological variables  \cite{paniconi2015physically,fatichi2016overview}, and lake temperature \cite{gaudard2019toward,hipsey2019general,winslow2017large}. Process-based models encode our understanding of relevant physical processes into numerical formulations. These relationships are often developed from decades of theory, observation, and experimentation, resulting in sufficient understanding of processes and their interactions to support defining them with code for a simulation model \cite{cuddington2013process, white2019should}. However, these models provide an approximation of reality and often require time-intensive parameter calibration to compensate for incomplete inclusion or resolution of processes. More recently, the rapid growth of sensor data \cite{porter2012staying, hampton2013big} along with advances in computation have led to development and increased use of powerful data-driven environmental models. Ensemble tree methods like gradient boosting and random forests, in addition to more advanced methods like deep learning \cite{goodfellow2016deep}, have been effectively used for geoscientific applications \cite{reichstein2019deep} and water resources \cite{erdal2013advancing,shen2018transdisciplinary,tyralis2019brief}. A major reason for this success
is that ML models, given sufficient data, can discern  patterns and structure in problems where complexity prohibits explicit programming of a system’s exact physical nature. Given this ability to automatically extract complex relationships from data, ML models (e.g., deep learning) appear promising for scientific problems with physical processes that are not fully understood by researchers, but for which data of adequate quality and quantity is available.  Given enough data, data-driven models can  increase prediction accuracy relative to existing process-based methods due to lack of \textit{a priori} constraints and the expressive power of modern data-driven models, though they can lack interpretability and generalizability, and they often fail to leverage domain knowledge. Coupling deep learning in particular with process-based models is an emerging paradigm for modeling earth systems, enabling the discovery of patterns that are not only generalizable but also consistent with existing scientific knowledge \cite{karpatne2018machine,shen2018transdisciplinary,reichstein2019deep}. For example, in \cite{jia2019physics,read2019process}, typically data-hungry long short term memory deep learning models \cite{hochreiter1997long} are augmented with process-based knowledge to predict lake temperature dynamics more accurately than both the process-based model and the standard deep learning model. This class of method has been called "process-guided deep learning" (PGDL\add{; see Table \protect\ref{tab:abbrev} for list of abbreviations}) and is an accelerating field of study \cite{willard2020integrating,kashinath2021physics}. Previous works modeling lake temperature at a broad scale have focused on calibrating parameters with available data, when data are unavailable, using recommended default values based on field and laboratory studies \cite{read2014simulating,winslow2017large}. These approaches have since been outperformed by PGDL models in cases of both high and low data availability \cite{read2019process}. However, in the case of no available temperature measurements to train or calibrate a model, no effort has yet been made to transfer PGDL models from well-\change{observed}{monitored} systems for prediction. 


Lakes are an exemplar for the disparity in observations across systems, where \textgreater80\% of \add{in-situ} water quality observations come from 20\% of \change{observed}{monitored} lakes \cite{stanley2019biases}, and \change{many more lakes are completely unobserved}{the majority of lakes have no in-situ monitoring data. In this work, we designate ``monitored" vs ``unmonitored" status of lakes based on the presence of in-situ data, and consider remote sensing integration in the discussion section}. How can we leverage the information in a small population of lakes to make predictions in the much larger population of sparsely \change{observed}{monitored} to completely \change{unobserved}{unmonitored} systems? First, temporal synchrony in characteristics across ecosystems suggests that information or models from a highly monitored system could be transferred to a less- or un-monitored system. Examples include synchrony between stream temperature and streamflow, between organic matter concentrations across different lakes \cite{baines2000synchronous,erlandsson2008thirty}, or coherence in lake temperature patterns \cite{benson2000regional}. Synchrony can emerge for a variety of reasons, including but not limited to shared underlying physical processes, weather conditions, or landscape context; patterns in synchrony across ecosystems therefore exhibit strong relationships to other physical variables. For instance, lake morphometric factors like maximum depth and surface area have a direct relationship to the stratification dynamics of lakes \cite{gorham1989influence, stefan1996simulated} and correlate with temporal coherence between lakes \cite{magnuson1990temporal, george2000factors}. Water clarity can also affect the responses of below-surface phenomena to solar radiation across different systems \cite{read2013physical, rose2016climate}. Differences in coherence strength can also be attributed to different dominant external drivers \cite{livingstone2008change}. Fortunately, many of these physical characteristics like shape, depth, and water clarity are more widely available than other measures of water quality. Further, these characteristics mediate the relationship between external drivers and within-lake responses (e.g., through sedimentation rates and head storage), such that information gained about dynamics in one lake could be transferred to other similar lakes, regardless of whether they exhibit temporal synchrony. Determining the generalizability of the relationship between physical characteristics and water quality dynamics across different ecosystems could allow the strategic transfer of site-specific models from \change{well-observed}{well-monitored} systems to predict temporal patterns in \change{unobserved}{unmonitored} systems.  




Currently, methods to extend accurate site-specific models to broad scale predictions are rare or nonexistent. In hydrology, extending site-specific parameterizations has been achieved through regionalization and catchment classification \cite{sivapalan2003iahs,wagener2007catchment,samaniego2010multiscale}. For example, \cite{samaniego2010multiscale} focus on transfer functions connecting geophysical attributes to process model parameters. However, these approaches are not widely regarded as successful, with noted drawbacks of (1) uncertainty in geophysical attributes, which translates to large uncertainty in parameter estimates, and (2) often-weak relationships between these attributes and parameters, perhaps because many of those parameters lack direct physical meaning \cite{archfield2015accelerating}. Water resources research has yet to establish a robust way to bridge scales for prediction accuracy for key ecosystem variables.

%


Transfer learning is a powerful technique for applying knowledge learned from one problem domain to another, typically to compensate for missing or nonexistent data in the new problem domain. The idea is to transfer knowledge from an auxiliary task, i.e., the source task, where sufficient labeled data is available, to a new but related task, i.e., the target task, often when data is scarce or inadequate \cite{pan2010survey, weiss2016survey}. Transfer learning using deep neural networks has shown recent success in ecological applications such as plant classification models \cite{kaya2019analysis}, air quality prediction \cite{ma2019improving}, and grassland fire risk assessment \cite{liu2019risk}. Transfer learning for deep neural networks is analogous to calibrating process-based models in \change{well-observed}{well-monitored} systems and transferring the calibrated parameters to models for \change{unobserved}{unmonitored} systems, which has shown success in hydrological applications \cite{kumar2013implications,roth2016model}. The task of deciding what model or parameters to transfer can be posed as a problem to be solved by meta-learning, or learning from previous learning experiences, which is another active area of machine learning research \cite{vanschoren2018meta,vanschoren2019meta}. In this paper, we focus on the meta-learning task of systematically learning how to map candidate source models (models trained on \change{well-observed}{well-monitored} lakes) to a particular task (prediction in \change{unobserved}{unmonitored} lakes) \cite{Brazdil2009}. For clarity, we define \textit{base-learning} models as the traditional machine learning models or process-based models that learn or are calibrated for specific tasks (e.g., prediction in a specific lake) as opposed to the \textit{meta-learning} model's goal of learning from a multitude of experiences transferring source models to target tasks. In the transfer learning context, which we call \textit{Meta Transfer Learning}, the meta-learning predicts which base models to transfer based on performance metrics for past transfer learning experiences and meta-features relating to the transferability of base-learning models \cite{ying2018transfer}. We demonstrate this method by transferring a suite of source lake temperature models to a number of artificially \change{unobserved}{unmonitored} target lakes, where temperature observations were only used for final evaluation. The metamodel was used to determine which source models would transfer well to the target lake and which lake attributes can best indicate the transfer performance.


Here, we demonstrate a meta transfer learning framework to predict lake temperature at depth. Our objectives are to (1) Demonstrate the use of a metamodel to rank both process-based models and process-guided deep learning models from \change{well-observed}{well-monitored} lakes (source lakes) in terms of their expected ability to predict lake temperature for a different, \change{unobserved}{unmonitored} lake (target lake); (2) Evaluate the MTL approach against existing process-based modeling approaches; and (3) Investigate the extent to which MTL can outperform the existing state-of-the-art process-guided deep learning models for the target lake itself in situations of limited observation data.

\section{Materials and Methods}
\subsection{Overview}
Here, we describe a method for model selection of trained source models from data-rich systems to predict lake water temperature in target systems with no data. The general idea of the MTL framework is visualized in Figure \ref{fig:meta-dia} and summarized as follows,

\begin{enumerate}
    \item Build and train two source models, a calibrated PB model and PGDL model, for each of the 145 \change{well-observed}{well-monitored} lakes.
    \item For each source lake, use all 144 other source models of the same type (PB or PGDL) to predict daily temperatures and evaluate prediction accuracy.
    \item Train the meta-learning model to predict the 145*144 collected model performance values from (2) based on \add{the} lake characteristics \add{that} we hypothesized could be important for selecting good transfer models.
    \item Given an artificially unmonitored target lake, where data is only used for final evaluation, and its meta-features, use the meta-learning model to predict model performance of each source model. Use the source model[s] with the lowest predicted error to model the target. 
\end{enumerate}

Sections \ref{method:pb-model} and \ref{method-pgdl} summarize the two types of source models, PB and PGDL models, respectively. Then, we describe the meta-learning model and how it is trained and used to identify lake features that predict successful transfers between source and target lakes \ref{method:mtl}. Lastly, Section \ref{method:data} describes the data used in Section \ref{method:experiment}, which contains descriptions of the experiments.

\subsection{Process-Based Models}
\label{method:pb-model}
As in previous studies of deep learning applied to lake temperature prediction \cite{read2019process,jia2019physics}, we chose the General Lake Model (GLM) 2.2 \cite{hipsey2019general} to represent process-based modeling, due to its proven ability to simulate thermal hydrodynamics in lakes along with its open‐source code availability (\url{https://github.com/AquaticEcoDynamics/GLM}). GLM can also be used to predict temperature at broad scales using widely-available lake characteristics (depth, surface area, clarity) to parameterize the model even when observations are not available \cite{winslow2017large}. \add{We acknowledge that GLM may not be the ideal process-based model in all calibrated and uncalibrated modeling scenarios, but consider the comparison of different process-based models for broad scale prediction to be out-of-scope of this work.}
Given that the MTL framework can use any similar hydrodynamic process-based model, we will further refer to the calibrated GLM using all the available observation data as ``PB" and the parameterized but uncalibrated version of GLM as ``PB0". To calibrate GLM for the PB models, we selected three parameters for calibration based on their known importance to model fits and their relative uncertainty: solar radiation scaling factor, momentum exchange coefficient, and hypolimnetic mixing efficiency. We used the \texttt{optim()} function in R \cite{team2013r} to modify these parameter values to minimize the RMSE of GLM temperature predictions relative to the available observations. See the Supplemental Information (text S3) in \citeA{read2019process} for details. 

\subsection{Process-Guided Deep Learning (PGDL) Models}
\label{method-pgdl}

We used a recently-developed PGDL model for lake temperature prediction,  \cite{jia2019physics,read2019process}, where process knowledge was combined with a Long Short Term Memory (LSTM) network via (1) a loss function term to encourage physical consistency and (2) pre-training using process-based model simulation data. \add{LSTM networks are part of a class of deep learning architectures built for sequential and time series modeling called recurrent neural networks \protect{\cite{hochreiter1997long}}.}\add{ These are particularly suited for predicting lake temperature dynamics given the often persistence of the response and the time lag between the input drivers and water temperature changes that can be represented in the memory properties of LSTM \protect\cite{jia2019physics,read2019process}.} Here, the simulation data used for pre-training are the output of the parameterized but uncalibrated version of the PB model (PB0) described in Section \ref{method:pb-model}. The components of the PGDL model are described in more detail in Supplemental Information (Text S1). The input features for the model \remove{(Table 1)} are the meteorological factors that contribute to incoming and outgoing heat fluxes and the depth (distance from surface) of the target prediction \cite{wetzel2000heat, fink2014heat, zhong2016recent}. \add{This includes short-wave and long-wave radiation (in W/m\textsuperscript{2}), air temperature (in \degree C), relative humidity (0-100\%), wind speed (in m/s), rain (in m/day), and snow (in m/day).} The meteorological features are identical to the drivers used in the GLM simulations except that they are each normalized to a mean of 0 and standard deviation of 1 based on a calculated global mean for each driver across all lakes\add{, a recommended step for training neural networks to address differences in the scales across input variables \protect\cite{sola1997importance}.}

\note{Table with Input Drivers Removed}


\subsection{Meta Transfer Learning with Gradient Boosting Regression}
\label{method:mtl}
\begin{figure}
  \centering
    \includegraphics[width=0.6\textwidth]{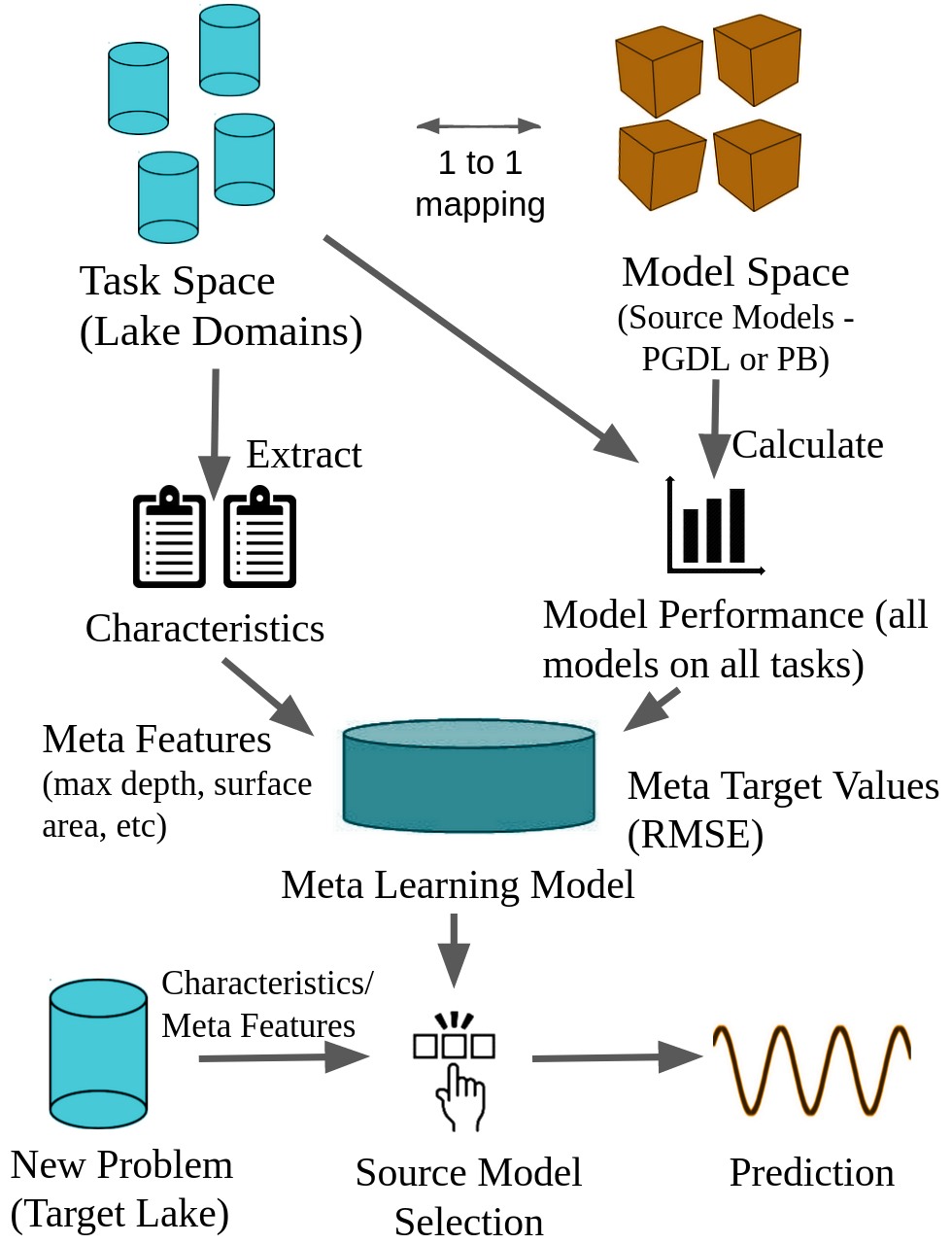}
  \caption{Meta-learning general framework\add{. The meta learning model (metamodel) is trained to predict source model performance (root mean square error; RMSE) based on lake domain characteristics (meta features). The performances and characteristics from all source models applied to all other source lakes are used for metamodel training. This trained model is used to predict source model performance and inform source model selection for a \textit{new} target lake. }}
  \label{fig:meta-dia}

\end{figure}

Our MTL framework aims to predict the accuracy of each source model on an \change{unobserved}{unmonitored} target lake. Here, two metamodels were built, one for predicting the performance of source PB models on target lakes (PB-MTL), and one for predicting the performance of source PGDL models on the same target lakes (PGDL-MTL). As shown in Figure \ref{fig:meta-dia}, each meta-learning model takes in lake-level features that may contain information about the transferability from a source to a target. We call these predictors \textit{meta-features}; meta-features included differences in physical attributes between the source and target lake, measures of data quality in the source lake, and features of the source and target that were derived from PB0 such as the likelihood of stratification. The response variable was the prediction accuracy (measured as root mean squared error, RMSEs) of transferring the source model (either PGDL or PB) to the target lake, where lower RMSEs represent a successful transfer between lakes. If $i$ is the index for the source lake, and $j$ the index of the target lake, the meta-features for each unique source-target pair can be written as $X_{\text{i}\to \text{j}}$, and target RMSE values as $RMSE_{\text{i}\to \text{j}}$. The function $\mathcal{F}$ we are attempting to approximate can then be written as

\begin{equation}
    \label{eq:metafunc}
    \mathcal{F}(X_{\text{i}\to \text{j}}) = RMSE_{\text{i}\to \text{j}}
\end{equation}

The training dataset for each metamodeling scenario then contains all $(n)*(n-1)$ possible source-target pairs as follows:

\begin{equation}
    \label{eq:metadata}
    \{(X_{i \to j}, RMSE_{i \to j})  |  i \neq j\}
\end{equation}

The following subsections describe details of this MTL approach, including the method of gradient boosting regression used for the metamodel, how meta-features were selected, and how gradient boosting hyperparameters were tuned.

\subsubsection{Gradient Boosting Regression}
Due to its predictive power, ease of implementation, and ability to illustrate the relationships between predictors and the response, we chose gradient boosting regression to predict the RMSE of source-target pairs from meta-features. In short, gradient boosting creates an ensemble of estimator models. It starts by fitting an initial regression tree model to the data. Regression decision trees are generated such that each decision node in the tree contains a test on the input variable's value, and the tree terminates with nodes that contain the predicted output variable values (RMSE in this case). Then, it builds a second model that prioritizes accurately predicting the cases where the first model performs poorly, a process known as boosting. The ensemble of these two models can be expected to perform better than the first model due to this new prioritization. Estimators are then continuously added until a set amount is reached. Gradient boosting in particular generalizes boosting by optimizing with a differentiable loss function, which in the case of regression is usually mean squared error (MSE). Further method details can be found in \citeA{friedman2001greedy}.

\subsubsection{Identification of Meta-Features}
\label{method:metafeature_id}
We started with a collection of candidate meta-features that we hypothesized could predict the performance of source models in predicting temperature in different target lakes. As in Equations \ref{eq:metafunc} and \ref{eq:metadata}, each set of meta-features ($X_{\text{i} \to \text{j}}$) is unique to a source-target lake pair. An exhaustive list of 96 possible meta-features is listed in Supplemental Information Table S1, which are divided into four categories: lake attributes, PB0 simulation statistics, general observation statistics, and meteorological statistics. The last two categories are commonly-used meta-features that either (1) use statistics relating to the quality and quantity of observations of the source lake, or (2) compare differences in the data distributions of input features between source and target domains  \cite{castiello2005meta}. We expand on this with the  two additional categories, lake attributes and a PB0 simulation statistic. All differences are calculated as the source value minus the target value.

\begin{enumerate}
    \item \textbf{Lake Attributes}: These features contain information about maximum depth, surface area, and other lake properties that are not directly used in model training since they are not features or observations, but may mediate or contain useful information about the physical response of lakes to meteorological drivers. They are calculated as the difference between the source and the target lake values. 
    \item \textbf{PB0 Simulation Statistic}: This feature describes an important property of the PB0 temperature predictions, the percentage of dates on which each lake is stratifies. We used PB0 predictions as a surrogate for in-situ temperature observations, which are not available for target lakes. The PB0 model translates driver data into temperature predictions via process understanding, and it can therefore give insight into similarities across lakes such as the likelihood the lake is stratified or how the lake responds to wind events. This statistic was already available as part of the pre-training process for PGDL models in this study, and is also calculated as a difference between the source and the target lake.
    \item \textbf{General Observation Statistics}: These features contain information about temperature measurements that only pertains to the source lake. Ideally they would contain information about the quality of the source data. For example, a very poorly \change{observed}{monitored} source lake without adequate data to train a model could indicate poor transfer performance. Example statistics include total observations, number of observations per season, mean depth where temperature was measured, and mean temperature measured.
    \item \textbf{Meteorological Statistics}: These features contain differences between the source and the target lake  in both annual and seasonal averages and standard  deviations of the 7 meteorological drivers used as inputs to the source models \remove{(Table 1)}. Examples include differences in mean air temperature, solar radiation, and relative humidity.
\end{enumerate}

Then, to narrow down the number of meta-features, we performed recursive feature elimination with cross validation (RFECV) \cite{guyon2002gene}. Recursive feature elimination is a feature selection method that fits a model and iteratively removes the weakest features until an ideal set that produces the lowest cross-validation error is reached. To do this we used two Scikit-learn python modules \cite{buitinck2013api}. For building the base model we used \texttt{GradientBoostingRegressor} with default parameters and 3000 estimators, and for performing feature selection we used the \texttt{RFECV} library with 24-fold cross validation and mean squared error loss. We also used the importance of each meta-feature to interpret how the transfers were selected. Here, feature importance was calculated by the \texttt{GradientBoostingRegressor} as a measure of how each feature affected mean squared error across nodes in the decision trees, weighted by how often those nodes are reached \cite{buitinck2013api}.

\subsubsection{Hyperparameter Tuning}
\label{method:hypertune}
For both PB-MTL and PGDL-MTL, we tuned two gradient boosting hyperparameters that are known to affect performance: the number of boosted decision trees and the learning rate (impact of each tree on final outcome). The remaining parameters were left at their default values for the  \textit{GradientBoostingRegressor} class in scikit-learn version 0.22.1. We construct a nested 24-fold cross validation (CV) to estimate the generalization ability of the model given certain hyperparameter values. This CV works by performing 24 iterations of removing 1/24th of samples from the dataset for validation and taking the average mean squared error as an estimate of model performance for a given set of hyperparameters. CV is done for every set of candidate hyperparameter values in an exhaustive search  of two candidate learning rates \{0.05, 0.10\} and intervals of 100 decision tree estimators from 1000 to 6000. The ideal hyperparameters were found to be learning rates equal to 0.05 for both PB-MTL and PGDL-MTL, and number of decision trees equal to 4500 for PB-MTL and 4900 for PGDL-MTL.

\subsection{Data}
\label{method:data}
\begin{figure}
    \centering
    \includegraphics[width=.7\textwidth]{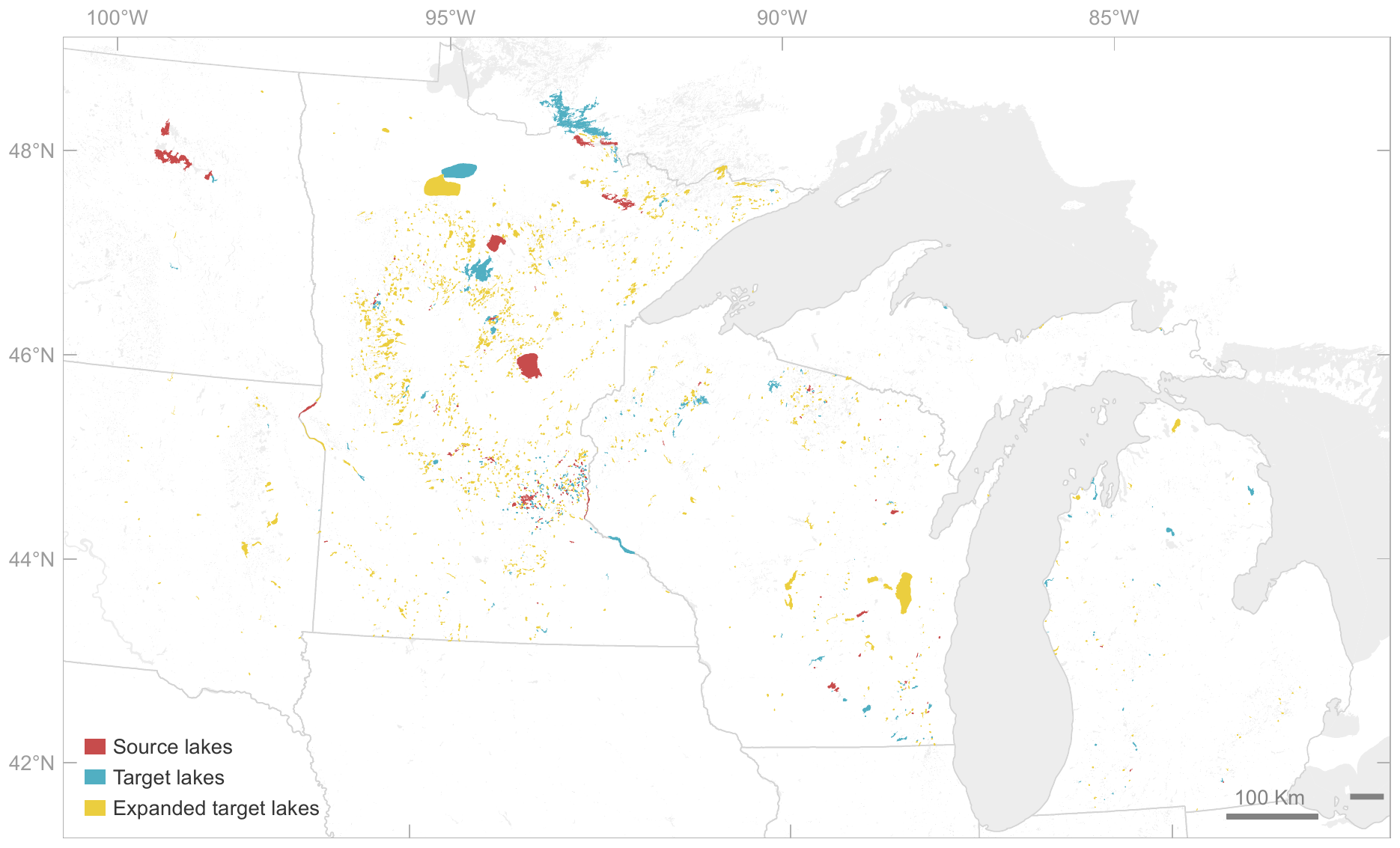}
    \caption{\change{Map of all lakes used in experiments. 145 source lakes are shown in red, 305 initial target lakes along with the additional 1882 expanded target lakes are shown in dark and light blue respectively.}{Map of all lakes used in experiments. 145 source lakes are shown in red, 305 initial target lakes are shown in blue, and the additional 1882 expanded target lakes are shown in yelllow.}}
    \label{fig:map}
\end{figure}

All the data used in this work is available through a data release on the U.S. Geological Survey's ScienceBase platform \cite{willard2020data}. All study lakes are located in the Midwestern United States, and details about the selected lakes are included in the data release. Briefly, 450 lakes met our data density criterion of at least 50 unique observation dates where there are at least one measurement for every two meters of depth or at least 5 total observations. From these lakes, in-situ lake temperature measurements between 1980 and 2019 were used to train and test all our models. To build the metamodel 145 of these lakes were used, and the rest are considered “artificially \change{unobserved}{unmonitored}”, where data is only used for final evaluation. An additional 1882 lakes with fewer observations were used as targets in an expansion exercise described in the Discussion (Figure~\ref{fig:map}).

\note{Figure~\protect\ref{fig:map} updated with changed color scheme and projection}

Meteorological data used as the input drivers for our models were gathered from the North American Lake Data Assimilation System (NLDAS-2) \cite{mitchell2004multi}. As in \cite{winslow2017large, read2019process}, these gridded data were transformed into process-model ready input (see \cite{hipsey2019general}). These inputs were then normalized for use in the machine learning models as mentioned in Section \ref{method-pgdl}. Lake attributes used as meta-features in the MTL algorithm were acquired in the same manner as in previous modeling studies in the region \cite{read2014simulating, read2019process}. A more detailed description of the sources and processing of these attribute data can be found in \cite{winslow2017large}.

\subsection{Model Experiments}
\label{method:experiment}

We designed two experiments that use the previously described metamodel built using meta-features and past model transfer RMSE.  For both experiments, we used 145 of the 450 \change{highly observed}{well-monitored} lakes as detailed in Section \ref{method:data} as source lakes, and we kept the remaining 305 lakes as target lakes for which the metamodel was used to select one of the 145 source lakes. Source lakes were selected to be representatively distributed across maximum depth values and log-scale surface area values (see Supplemental Information Figure S3). In all experiments the metamodel training data consisted of RMSEs from applying each of the 145 source lake models on all other source lakes, leading to 144*145 meta-learning data points (20880 total). Then, after the metamodel is trained, for each source-target pair we constructed the meta-features as described in Section \ref{method:metafeature_id}. From these meta-features, both metamodels (PB-MTL, PGDL-MTL) were then used to predict the expected RMSE of each of the 145 source models when transferred to the target lake.

\subsubsection{Experiment 1: Predicting Temperatures in ``Unmonitored" Lakes}
\label{method:exp1}
Experiment 1 evaluates the performance of the meta transfer learning models in a real-world scenario: predicting water temperature at multiple depths in unmonitored lakes. Given the predictions of source model performances from both metamodels, the PB or PGDL model with the lowest predicted source-to-target RMSE was singled out for use on each target lake. We compared the top-predicted transfers for each of the 305 test lakes against its PB0 simulation. 

We assumed that our metamodel would not be able to select the true best source lake in all instances. We therefore also evaluated an ensemble method that combined several of the top predicted models. Ensembles of multiple individual models that perform well can almost always improve over their average prediction error\cite{zenobi2001using,kuncheva2003measures}. This was proven by \citeA{krogh1995neural}, who showed that increasing ensemble diversity (the extent to which single models disagree), given constant average error of individual ensemble members, reduces the overall ensemble error. This reduction often occurs because some ensemble member predictions are biased positively and some negatively, leading to bias cancellation in the ensemble prediction. We used a simple ensemble that takes an unweighted average of the predictions from each selected PGDL source model for each date and depth. Lakes selected for ensembling were the top ``$n$" source models predicted to have the lowest RMSE on the test lake. The optimal value of $n$ was estimated using a 29-fold cross validation. In each cross validation fold, 5/145 source lakes were designated as validation lakes and a metamodel with the same hyperparameters as described in Section \ref{method:hypertune} was trained on the remaining 140 lakes. Then, $n$ source lakes were selected for each validation lake. Estimated ensemble error was then the mean ensemble errors across all folds. This was repeated for values of $n$ ranging from 2 to 10, where 9 was found to be the optimum, but values differed minimally between 5 and 10. We call this 9 source ensemble approach PGDL-MTL9. 

Lastly, we examined the metamodels themselves. In addition to evaluating the performance of the predicted best source-target transfer, we looked at how well the metamodel predicted the RMSE of \textit{every} source-to-target combination and how well it was able to rank the source models. For the former we calculate a median across all target lakes of the metamodels' predictions for RMSEs and the actual source-to-target RMSEs. For the ranking evaluation, we used the Spearman rank correlation coefficient shown as $r_s$. We also looked at distributions of the actual ranks (for the RMSEs of sources actually applied to targets) of those models that were identified by the metamodel as the top or top 9.

\subsubsection{Experiment 2: Comparing PGDL-MTL with PGDL for Sparsely \change{Observed}{Monitored} Systems}
\label{method:exp2}
Experiment 2 examines the extent to which PGDL-MTL is an improvement over PGDL in systems that have some observations but are not sufficiently \change{observed}{monitored} to train any traditional deep learning model effectively. \add{In this experiment, we define ``sparsely monitored" as between 1 and 50 sampling dates.} Deep learning models are generally data-hungry, but PGDL models pre-trained on PB0 output have shown to achieve high accuracy with only a few observations \cite{jia2019physics,read2019process}. Thus, both PGDL and PGDL-MTL  have the potential to alleviate the difficulty in calibrating process-based models for sparsely \change{observed}{monitored} lakes, where overfitting can be a problem. However, PGDL-MTL has the potential to harness more data and thereby outperform PGDL. The situation of few available observations is also far more common than the \change{well observed}{well-monitored} case of the lakes chosen in this work \cite{read2017water}. To that end, we artificially sparsified the data available in the 305 test lakes to train PGDL models on low amounts of data. Then, these low-data PGDL models were compared to the PG-MTL and PGDL-MTL results of Experiment 1. We used this comparison to estimate the data threshold where PGDL tends to outperform PGDL-MTL.

Artificial sparsity was induced by building five PGDL models for each suitable lake for twelve different amounts of sampling dates (1, 2, 5, 10, 15, 20, 25, 30, 35, 40, 45, 50) used for training. The test period was set as the first third of temperature observations in time for each lake, leaving the training period as the last two thirds of temperature observations. For each sampling date treatment we used all lakes that had at least that number of sampling dates during the training period. Of the 305 possible lakes, 221 had 50 or more observations, 270 had 40 or more observations, and all 305 had 30 or more observations. For the five models within each lake and data availability category, variability was introduced by randomly selecting dates for the training data. 


\section{Results}



\begin{figure}[h]
    \centering
    \includegraphics[width=\textwidth]{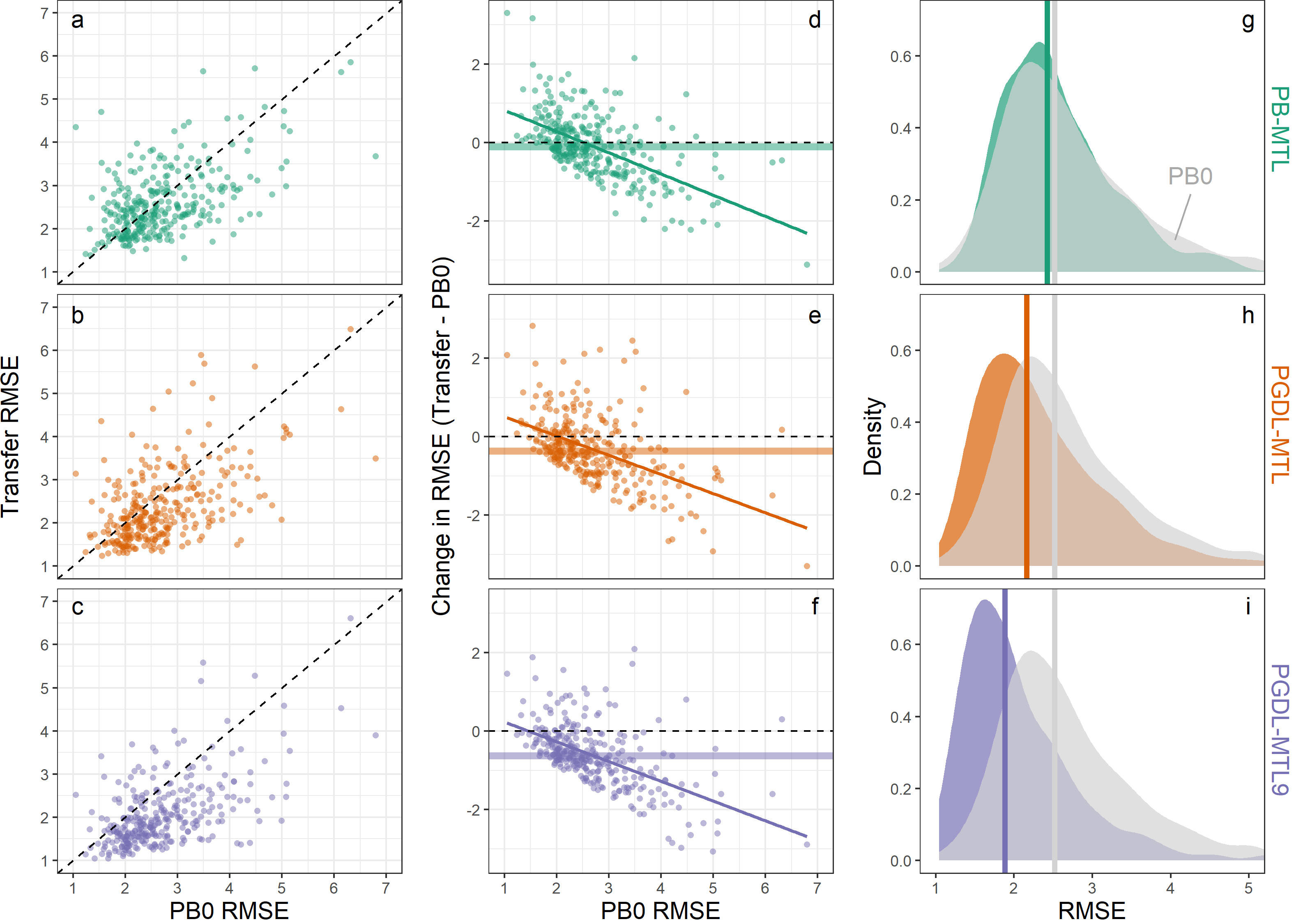}
    \caption{Comparison of the performance of the three MTL approaches relative to PB0 on 305 lakes. \change{PB-MTL and PGDL-MTL are the transfer of process-based and PGDL models respectively, and PGDL-MTL9 is an averaged ensemble prediction of the top 9 PGDL models.}{a-c) RMSE of PB0 relative to the three transfer models, where the dotted line shows the 1:1 relationship. d-f) The difference between RMSE of the transfer and PB0 models, where the black dotted line shows the zero or no change line, and the solid colored lines show the linear regression fit of the change in RMSE as a function of PB0 RMSE. g-i) The distribution of RMSE from PB0 and transfer models, where the vertical gray and colored lines are the median PB0 and transfer RMSE, respectively. PB-MTL (a,d,g) and PGDL-MTL (b,e,h) are the transfer of process-based and PGDL models respectively, and PGDL-MTL9 (c,f,i) is an averaged ensemble prediction of the top 9 PGDL models.} }
    \label{fig:305accuracy}
\end{figure}
\begin{table}[h]
\caption{\textit{Results of PB-MTL and PGDL-MTL Applied to Test Lakes.}}
\centering
\label{tab:305accuracy}
\begin{threeparttable}
\scalebox{0.78}{
\begin{tabular}{ |l|c|c|c|c|c|c| } 
\toprule
 Method & \thead{Median RMSE\\ (\degree C)} & \thead{Lower \\quartile RMSE} & \thead{Upper \\quartile RMSE} &\thead{Median\\meta RMSE} & Median $r_s$\\
\midrule
PB0 & 2.52 & 2.07 & 3.12 & --&--\\ 
\midrule
PB-MTL & 2.42 & 2.04 & 2.95 & 0.853 & 0.653\\
\midrule
PGDL-MTL & 2.16 & 1.74 & 2.81 & 0.871 & 0.663\\ 
\bottomrule

\end{tabular}}
\begin{tablenotes}
\item \textit{Note}. The first three columns are the quartile distributions of RMSE of the best predicted source lake for each test lake. The fourth column is the median RMSE between the metamodel-predicted RMSEs and the observed RMSEs. The fifth column is the median Spearman rank correlation coefficient between the metamodel-predicted RMSEs and the actual RMSEs.
\end{tablenotes}
\end{threeparttable}
\end{table}

{\parindent0pt\paragraph*{PB- and PGDL-MTL model accuracy on 305 test lakes}\hfill}

In Experiment 1, PGDL-MTL and PB-MTL predictions of water temperature in the 305 test lakes were typically more accurate than predictions from the uncalibrated process-based model (PB0; Table \ref{tab:305accuracy} and Figure \ref{fig:305accuracy}). The median RMSE across the test lakes was 2.42\textdegree C for PB-MTL and 2.16\textdegree C for PGDL-MTL, versus 2.52\textdegree C for PB0. PB-MTL outperformed PB0 for 203/305 of the lakes and PGDL-MTL outperformed PB0 for 226/305 of the lakes, and the amount of improvement the transfer provided generally increased with PB0 error (Figure \ref{fig:305accuracy}). \add{Predictions of deeper water temperatures from the transferred models had higher RMSEs in general as compared to the lake-specific accuracy of all depths, with the highest median RMSE from PB0 models, followed by PB-MTL, and with PGDL-MTL having the lowest median deep-water RMSE (2.59\textdegree C, 2.56\textdegree C, and 2.36\textdegree C, respectively; RMSEs calculated based on predicted versus observed temperatures at or below 75\% of the maximum depth of the lake; 18 of 305 lakes had no observations at these depths)}.  

\begin{table}[h]
    \centering
    \caption{\textit{Median Actual RMSE of PGDL Source Models of Different Metamodel-Predicted Ranks.}}
    \scalebox{0.8}{
    \begin{tabular}{|c|c|c|c|}
    \toprule
    Source system(s) & \thead{Median RMSE\\ (\textdegree C)}  & \thead{Lower\\quartile RMSE} & \thead{Upper\\quartile RMSE} \\
    \midrule
        Rank 1 Source  & 2.16 & 1.73 & 2.80\\
    \midrule
        Rank 2 Source  & 2.21 & 1.79 & 2.77 \\
                 \midrule
        Rank 3 Source& 2.15 & 1.75 & 2.82\\
                 \midrule
        Rank 4 Source& 2.20 & 1.85 & 2.86\\
                 \midrule
        Rank 5 Source& 2.20 & 1.78 & 2.83\\
                 \midrule
        Rank 6 Source& 2.25 & 1.85 & 2.86\\
                 \midrule
        Rank 7 Source& 2.23 & 1.84 & 2.86\\
                 \midrule
        Rank 8 Source& 2.24 & 1.84 & 2.90\\
                 \midrule
        Rank 9 Source& 2.21 & 1.83 & 2.90\\
             \midrule
        9 Source Ensemble & 1.88 & 1.56 & 2.41\\
         \bottomrule
    \end{tabular}}
    \label{tab:err_per_source}
\end{table}

{\parindent0pt\paragraph*{Model ensemble performance}\hfill}

Additionally, the ensemble PGDL-MTL9 model provided still better performance than PGDL-MTL. We can see in Table \ref{tab:err_per_source} that the RMSE of the combined averaged prediction of the source models tended to be lower than most of the source models individually. The ensemble model PGDL-MTL9 had a median RMSE of 1.88 \textdegree C, which is an improvement over the single-source PGDL-MTL, which had a 2.16 \textdegree C median RMSE. When compared to PB0 in Figure \ref{fig:305accuracy}, PGDL-MTL9 outperforms PB0 for 260/305 of the test lakes.  Table \ref{tab:err_per_source} also shows the distribution of RMSE values per source systems at given ranks, between 1 and 9, as predicted by the metamodel.  Comparing the individual source model RMSEs across the top 9 ranks, we see ranges of only 0.09 \textdegree C  in median RMSE, 0.12 \textdegree C in lower quartile RMSE, and 0.13 in upper quartile RMSE. 

{\parindent0pt\paragraph*{Meta-features and importances}\hfill}

The top selected meta-features were related to maximum depth in both PB-MTL and PGDL-MTL, with combined importances of 50\% and 45\%, respectively. Surface area, observation count, source lake observed temperature, and stratification indicators were selected as meta-features in both PB-MTL and PGDL-MTL but were of lesser importance (Table \ref{tab:feat_important}). 

\begin{table}[h]
  \caption{\textit{Selected Features for PB-MTL and PGDL-MTL and  Importances}}
  \centering
  \scalebox{0.87}{
    \begin{tabular}{|l|l|l|}
    \toprule
    \textbf{Meta-feature} & \multicolumn{2}{c|}{\textbf{MTL importance}} \\
    \cmidrule{2-3}
                 & PB & PGDL\\
    \midrule
    Max Depth Difference & 0.39 & 0.26\\
    Max Depth Percent Difference & 0.11 & 0.19 \\
    GLM Stratification Percent Difference & 0.18 & 0.066\\
    Surface Area Difference & 0.065 & 0.087\\
    Surface Area Percent Difference & 0.037 & 0.087\\
    Mean Source Observation Temperature & 0.037 & 0.072\\
    Number of Source Temperature Observations & 0.028 & 0.072\\
    Square Root Surface Area Percent Difference& \NA & 0.085 \\
    Lathrop Stratification Difference& 0.020 & 0.034\\
    Autumn Relative Humidity Difference& \NA & 0.048 \\
    Source Observation Temp and Target Air Temp Difference & 0.033 & \NA \\
    Mean Autumn Wind Speed Difference & 0.027 & \NA\\
    GLM Stratification Absolute Difference & 0.024 & \NA\\
    Kurtosis Source Observation Temperature & 0.023 & \NA\\
    Mean Autumn Shortwave Difference & 0.020 & \NA \\
    Skew Source Observation Temperature & 0.017 & \NA\\
    \bottomrule
    \end{tabular}}
    \label{tab:feat_important}
\end{table}


\begin{figure}[h]
    \centering
    \includegraphics[width=\textwidth]{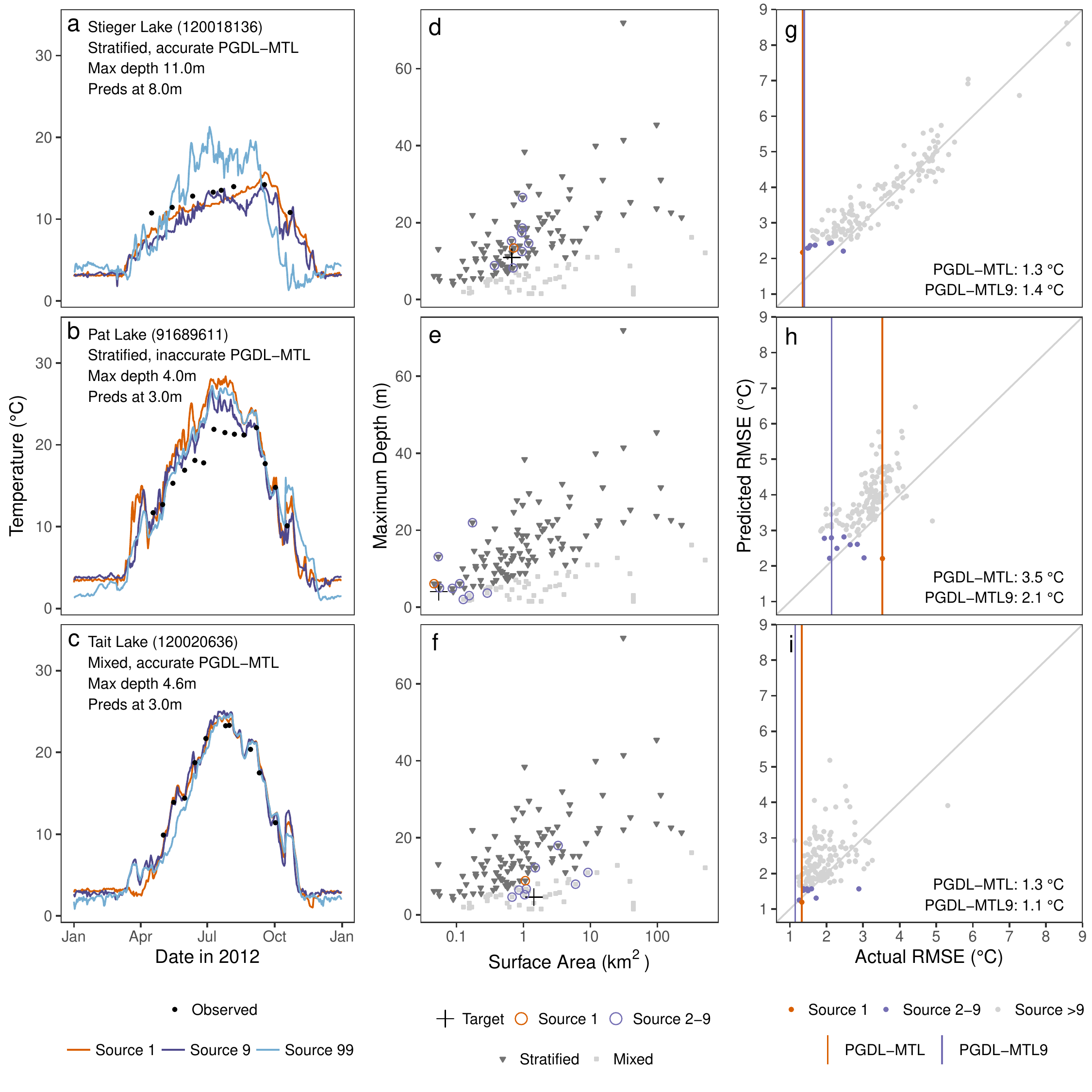}
    \caption{\protect\note{Replaced Figure 4. }\protect\change{Three example lakes to illustrate the application of PGDL-MTL and PGDL-MTL9. Panels a-c: Time series predictions at two depths in 2013-2014 for each target lake from the top-ranked PGDL source (Source 1), 9th-ranked source (Source 9), and a lower-ranked source (Source 99), with observed values (triangles) and PB0 predictions (gray lines) for comparison. Panel d: metamodel selections of source lakes for each lake, arranged by four features that dominated the MTL predictions: maximum depth (y axis), surface area (x axis), source lake observation count (circle size), and predicted stratification (darker = stratified). Panel e: Metamodel-predicted RMSEs versus actual RMSEs for the three example lakes.}{Deep-water predictions for three example lakes to illustrate the application of PGDL-MTL and PGDL-MTL9. Lakes were selected to represent successful and unsuccessful PGDL-MTL results for stratified lakes (rows 1 and 2, respectively) and the easier case of a mixed lake (row 3). Steiger, Pat, and Tait Lakes have 2,573, 469, and 3,865 total temperature observations, respectively. Panels a-c: Time series predictions at two depths in 2012 for each target lake from the top-ranked PGDL source (Source 1), 9th-ranked source (Source 9), and a lower-ranked source (Source 99), with observed values (points) for comparison. Panels d-f: metamodel selections of source lakes for each lake, arranged by three features that dominated the MTL predictions: maximum depth (y axis), surface area (x axis), and predicted stratification (darker = stratified). Panels g-i: Metamodel-predicted RMSEs versus actual RMSEs (for all depths and years) for the three example lakes.}}
    \label{fig:time_series}
\end{figure}

{\parindent0pt\paragraph*{Example time series prediction}
\hfill}

The metamodels typically chose source models that were good, but not optimal, matches to the target lake (Figure \ref{fig:time_series}). \remove{In three example lakes chosen to span the range of PGDL-MTL accuracies (RMSEs of 1.6 to 5.6 C), shallow-water predictions typically matched shallow-water observations (Figure \protect\ref{fig:time_series}}\remove{a-c). One stratified lake had stratified PGDL-MTL predictions (Figure \protect\ref{fig:time_series}}\remove{a) and another did not (Figure \protect\ref{fig:time_series}}\remove{b); the unstratified lake had unstratified predictions (Figure \protect\ref{fig:time_series}}\remove{c). Although the metamodel tended to select source lakes with similar maximum depths as their target lakes, some attention to surface area, the number of source observation dates, and the PB0-predicted stratification status was also evident (Figure \protect\ref{fig:time_series}}\remove{d).}\add{In a stratified lake with high PGDL-MTL accuracy (RMSE = 1.3 \textdegree{}C), top-ranked source models all came from stratified source lakes (Figure \protect\ref{fig:time_series}}\add{d) and captured the summer stratification dynamics (Figure \protect\ref{fig:time_series}}\add{a). In a stratified lake with low PGDL-MTL accuracy (RMSE = 3.5 \textdegree{}C), top-ranked source models came from a mix of stratified and unstratified source lakes (Figure \protect\ref{fig:time_series}}\add{e) and had similar predictions to a low-ranked source model (Figure \protect\ref{fig:time_series}}\add{b). In our 305-lake test set, mixed lakes (n=121) had lower mean RMSEs (mean=2.01, SD=0.52 \textdegree{}C) than stratified lakes (n=184; mean=2.62, SD=0.96 \textdegree{}C). Our mixed example lake illustrates that all candidate source lakes had lower RMSEs (Figure \protect\ref{fig:time_series}}\add{i) and similar predictions (Figure \protect\ref{fig:time_series}}\add{c) such that even though the metamodel selected a combination of mixed and stratified source lakes, the resulting RMSEs could still be quite low (PGDL-MTL: 1.3 \textdegree{}C, PGDL-MTL9: 1.1 \textdegree{}C). Consistent with the meta-feature importances in Table \protect\ref{tab:feat_important}}\add{, the selected source lakes tended to be similar to the target lake with respect to not just stratification but also maximum lake depth and surface area (Figure \protect\ref{fig:time_series}}\add{d-f). Ensembling with PGDL-MTL9 yielded similar accuracy to PGDL-MTL for the two example lakes with high PGDL-MTL accuracy (Figure}\add{ \protect\ref{fig:time_series}}\add{g,i) and substantially improved accuracy in the example lake where the PGDL-MTL model failed to capture the observed stratification dynamics (Figure \protect\ref{fig:time_series}}\add{h).}


\begin{figure}[h]
    \centering
    \includegraphics[width=\textwidth]{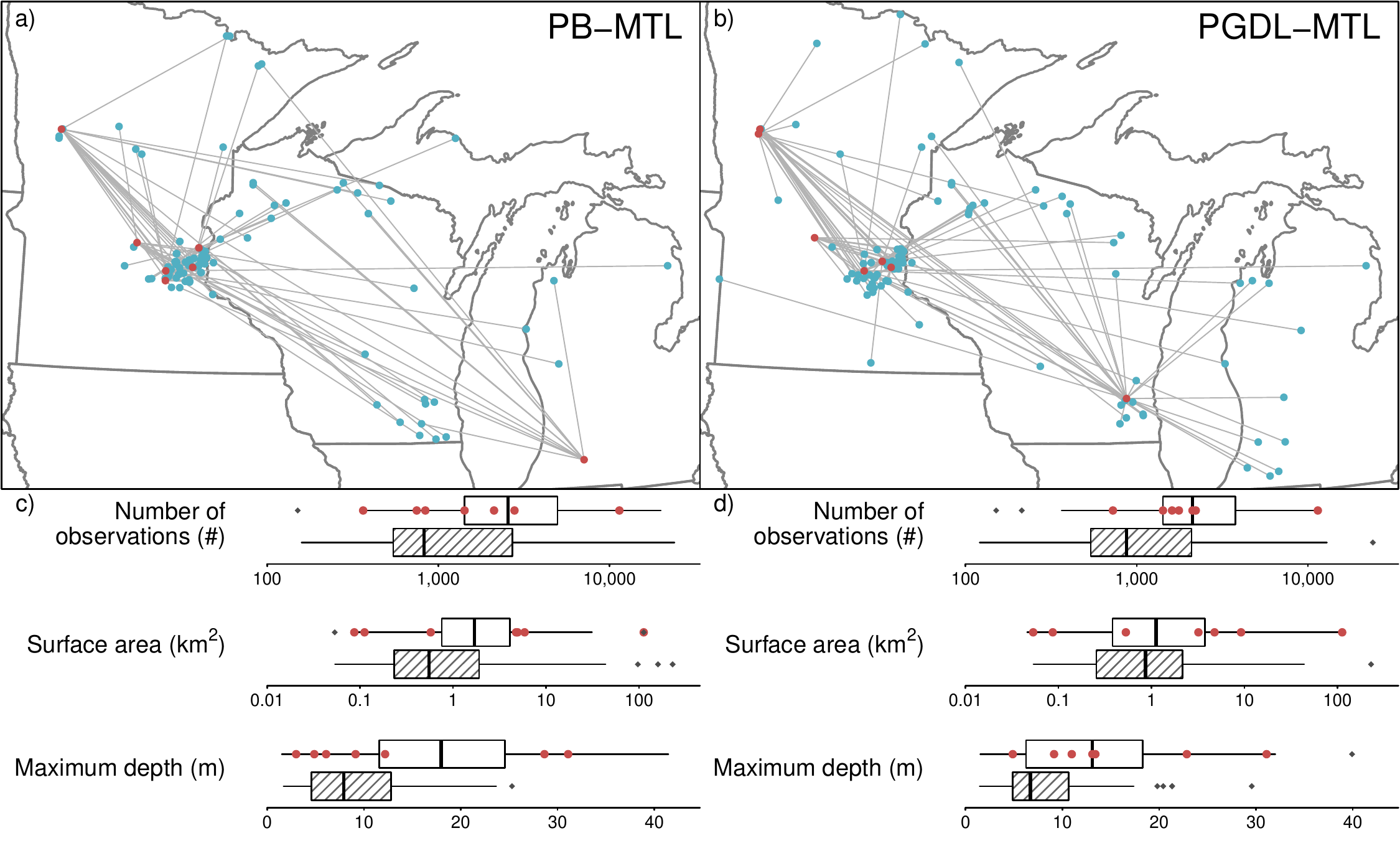}
    \caption{Top-selected source models compared to lesser-selected sources. In a), the seven process-based (PB) models chosen as a top source for ten or more target lakes by the meta transfer learning (MTL) model are shown in red, with grey lines connected to the paired target lake location; b) is the same as a) but for process-guided deep learning source models. In c), properties of lakes in the upper quartile of commonly chosen PB source models (white fill boxplot) are compared to the lowest quartile (hashed fill boxplot; based on MTL rank). Red dots represent the location of the seven source lakes featured in a). d) is the same as c), but for process-guided deep learning source models.}
    \label{fig:source_property_figure}
\end{figure}

{\parindent0pt\paragraph*{Features of best and worst source lakes}\hfill}

There were large differences in the frequency at which source models were chosen by the MTL to represent target lakes, and several factors emerged that suggested differences exist between commonly and rarely selected source lakes. A small fraction of source models were used to predict almost one third of target lake water temperatures, and eleven lakes were selected as the top PGDL or PB source for ten or more target lakes. Seven top PGDL source models were used for 100 target lakes and seven PB models for 95 of 305 target lakes, and three lakes were in this top category for both PGDL and PB models. In contrast, 59 PGDL and 64 PB source models were not chosen as a top model for any target lake (31 were never selected as sources in either model). Additionally, we summed the number of times each lake was predicted to be in the top 9 sources for the ensembles, and compared the raw lake attributes of the upper and lower quartiles (Figure \ref{fig:source_property_figure}). For both PGDL and PB transfer models, lakes that were transferred often were in general deeper, larger, and more \change{well observed}{monitored} than minimally transferred lakes. For PGDL, source models in the lower quartile of MTL selections had a median depth of 6.71 m, a median surface area of 0.86 km2, and a median of 872 training observations, compared to 13.1 m, 1.12 km2, and 2,117 observations for the upper quartile medians, respectively. The lower quartile of PB-MTL source models had medians of 7.9 m, 0.55 km2, and 824 calibration observations, with upper quartile medians of 18 m, 1.7 km2, and 2,557 calibration observations.


{\parindent0pt\paragraph*{Metamodel performance}\hfill}

\begin{figure}[h]
    \centering
    \includegraphics[width=\textwidth]{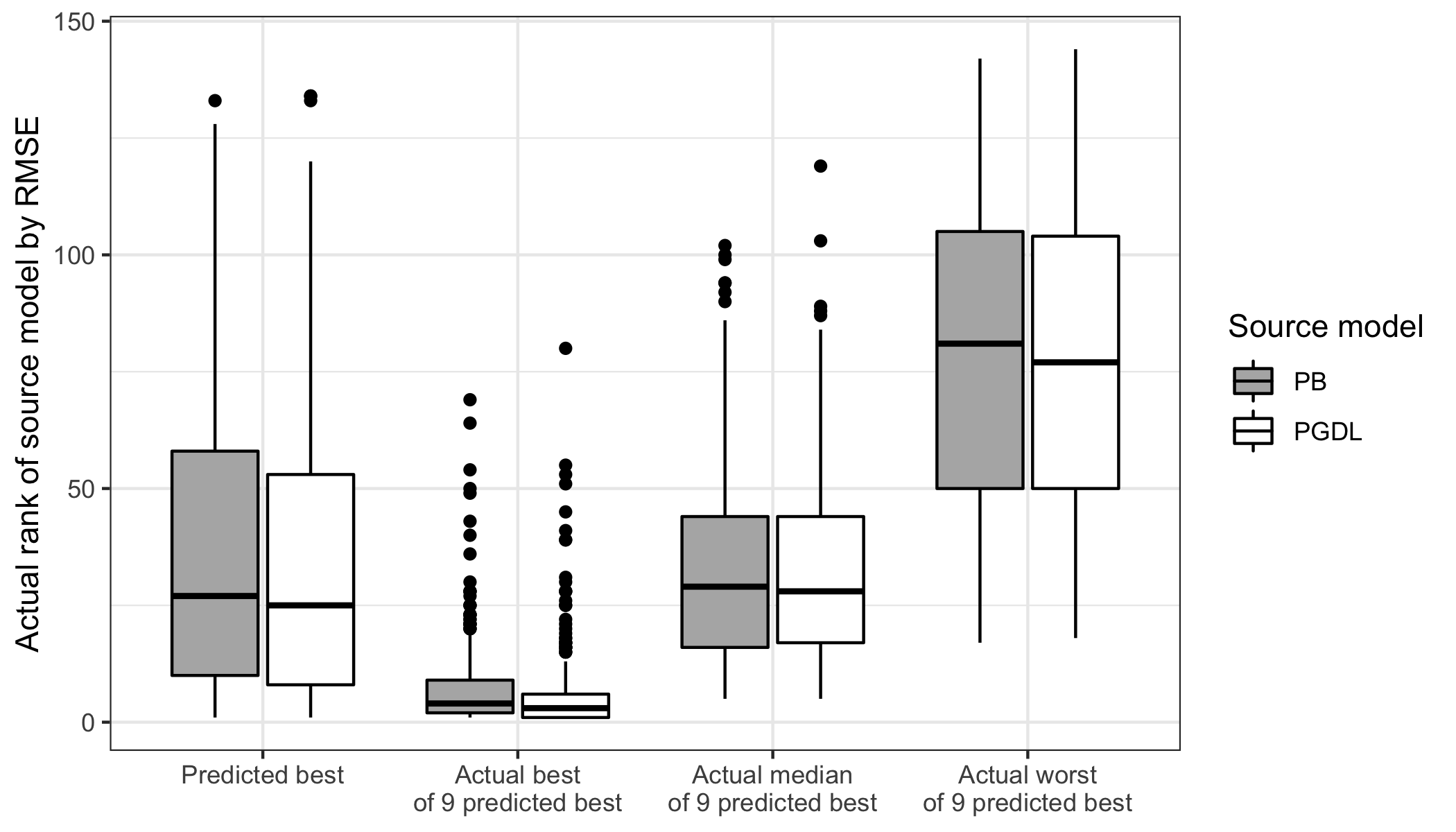}
    \caption{Plot showing the distribution of actual ranks of the metamodel-predicted top source models, for metamodels built on either PB sources (gray fill) or PGDL sources (white fill). Leftmost pair of bars: actual ranks for top-predicted models for each of the 305 target lakes. Other bars: best, median, and worst of the top-9-predicted sources.}
    \label{fig:meta_beanplot}
\end{figure}

To assess the metamodel's ability to predict the performance of source lake models, we looked at both the RMSE of the predicted RMSE versus the actual RMSE when transferring source models in Experiment 1, and also the ability of the metamodel to accurately rank source models from best to worst in the form of the Spearman rank correlation coefficient. The median meta-RMSE for PB-MTL was 0.853\textdegree C and the Spearman rank correlation coefficient $r_s$ was 0.659, and for PGDL-MTL the meta-RMSE was 0.871\textdegree C with an $r_s$ of 0.663 (Table \ref{tab:305accuracy}). Then, in Figure \ref{fig:meta_beanplot}, in addition to showing the distribution of actual ranks for the predicted best source PGDL model for each target system, we also show the distribution of ranks for sources within the 9 source ensemble PGDL-MTL9. Further visualization of the ranking ability of the metamodels is shown in Supplemental Information Figure S2. Here, we see that the two metamodels have similar predictive ability, with PGDL-MTL ranking slightly better as seen in the Spearman coefficient values.

\clearpage


{\parindent0pt\paragraph*{Comparison with data-sparse target lake models}\hfill}

\begin{figure}[h]
    \centering
    \includegraphics[width=\textwidth]{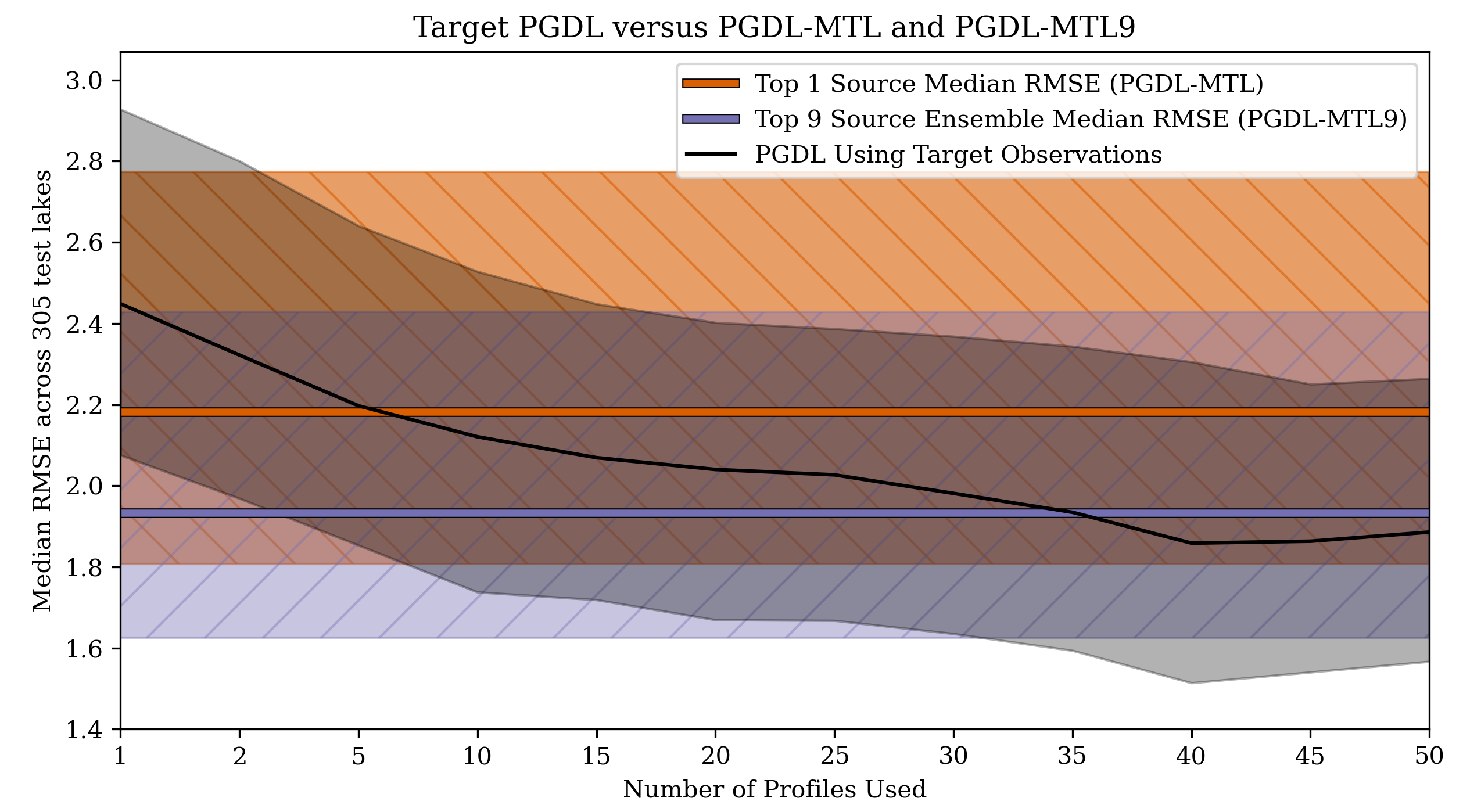}
    \caption{Median RMSE for PGDL trained with differing numbers of temperature profiles, with error bars representing upper and lower quartiles of the median RMSE across the 5 randomized selections of observations for each target lake. Colored horizontal lines represent the median RMSE with a band showing the range from lower to upper quartile for PGDL-MTL and PGDL-MTL9}
    \label{fig:exp2}
\end{figure}

In Experiment 2, the median RMSE across 305 test lakes \change{decreased monotonically}{tended to decrease} as the number of sampling dates used for training increased (Figure \ref{fig:exp2} and Table \ref{tab:pgdl_target}). Figure \ref{fig:exp2} shows performance of PGDL trained with differing numbers of temperature profiles compared to the MTL approach, and Table \ref{tab:pgdl_target} shows the specific RMSE numbers for Figure \ref{fig:exp2}. Here, the RMSE for each test lake is defined as the median RMSE across 5 randomly chosen sets of the same number of observations. Given that the RMSEs of the single-source PGDL-MTL and ensemble-of-sources PGDL-MTL9 from the previous experiment were 2.16 \textdegree C and 1.88 \textdegree C, respectively, PGDL models trained only on the target lake's data met or exceeded median MTL performance at between 5 and 15 observations for PGDL-MTL and between 35 and 40 observations for PGDL-MTL9. In other words, even for a reasonably \change{well observed}{well-monitored} lake (up to 40 observations), it can be better to borrow a model from a different and \change{better-observed}{better-monitored} lake than to train a model on the target lake observations. For context, 45 profiles is approximately the coverage a lake would have if it had a monitoring program that took a temperature profile monthly during the ice-free period for ~6 years.

\begin{table}[h]
    \centering
    \caption{\textit{Data for Figure \ref{fig:exp2}, Performance of PGDL Trained on Various Amounts of Target Lake Temperature Data Profiles}}
    \begin{threeparttable}
    
    \scalebox{0.8}{
    \begin{tabular}{|c|c|c|c|}
        \toprule
         \thead{Target\\Profiles} & \thead{Median of \\ Median RMSE \\(\textdegree C) } & \thead{Lower quartile\\ of Median RMSE \\(\textdegree C)} & \thead{Upper quartile\\ of Median RMSE \\(\textdegree C)}\\
         \midrule
         1 & 2.45 & 2.08 & 2.93\\
         \midrule
         2 & 2.32 & 1.97 & 2.80\\
         \midrule
         5 & 2.20 & 1.85 & 2.64\\
         \midrule
         10 & 2.12 & 1.74 & 2.53\\
         \midrule
         15 &  2.07 & 1.72 & 2.45\\
         \midrule
         20 & 2.04 & 1.67 & 2.40\\
         \midrule
         25 & 2.03 & 1.67 & 2.39\\
         \midrule
         30 & 1.98 & 1.64 & 2.37\\
         \midrule
         35 &  1.93 & 1.59 & 2.34\\
         \midrule
         40 & 1.86 & 1.51 & 2.31\\
         \midrule
         45 & 1.86 & 1.54 & 2.25 \\
        \midrule
         50 & 1.89 & 1.57 & 2.26 \\
         \bottomrule
        
    \end{tabular}}
    \begin{tablenotes}
    \item \textit{Note}. Medians of medians are calculated as the median across 305 test lakes of the median of 5 models trained with different random selections of observations.
    \end{tablenotes}
    \end{threeparttable}

    \label{tab:pgdl_target}
\end{table}

{\parindent0pt\paragraph*{Baseline performance of PB and PGDL source models}\hfill}

Success in transfer learning depends both on (1) metamodel success in choosing the best of the available source models for a target lake and (2) the baseline performance of the source models that could be transferred. If the PGDL-MTL metamodel had selected the best available source PGDL model for every target lake, the median RMSE would have been 1.54 \textdegree C, versus an RMSE of 1.79 \textdegree C if the best PB model was selected every time.  This difference, aligning with established knowledge that PGDL predicts more accurately than PB \cite{read2019process,jia2019physics}, can explain how the RMSE across the test lakes of PGDL-MTL was lower than PB-MTL even though PB-MTL had a lower meta-RMSE predicting the performance of source models.

\section{Discussion}
In this paper, we show Meta Transfer Learning (MTL) can be used to address monitoring gaps in environmental and ecological sciences by predicting in unmonitored systems. Even with the data deluge resulting from modern sensor developments, the majority of lakes and streams are unmonitored or have sparse observations. This has made it difficult to calibrate process-based models for these systems due to risk of overfitting, and even more inaccessible for traditional deep learning models which can require thousands or millions of data points. The MTL paradigm in this work harnesses data from many other systems to accurately predict temperature in unmonitored systems. Specifically, the transfer process leverages observations from \change{highly observed}{highly monitored} systems, simulated temperature data from process models, past model performance measures, and thousands of past transfer learning experiences to alleviate the drawbacks of both deep learning and process-model calibration in \change{unobserved}{unmonitored} systems. 

As experts in the water resources community have called for integration of process-based and data-driven methods \cite{hipsey2015predicting, shen2018transdisciplinary}, MTL involves a collection of approaches harnessing both ML and process knowledge. Here, we use the ML technique of gradient boosting regression for the meta-learning task of predicting the transferability of source models including those that employ Process-Guided Deep Learning (PGDL), which iteself integrates process knowledge into ML. Limnology domain expertise was also used in defining the candidate meta-features offered to the metamodel. The top selected meta-features matched our process understanding from dozens of studies that show relationships between the properties of lakes (surface area, depth) and physical responses to external drivers \cite{gorham1989influence, stefan1996simulated}. This work shows that these lake characteristics, which are more widely available than water quality data themselves, can be used to transfer information from highly \change{observed}{monitored} to \change{unobserved}{unmonitored} systems. 

Different types of lake-specific data were used to determine which sources should be transferred. Lake maximum depth difference between the source and target lake emerged as the most important in both the PGDL-MTL and PB-MTL approaches. Surface area differences were also included in both, but of less importance. This aligns with existing process-based lake modeling knowledge that maximum depth and surface area are key factors in lake stratification and thermodynamics \cite{gorham1989influence, stefan1996simulated}. Other  meta-features related to source model quality, like the number of observations and mean observation temperatures, were also included in both metamodels. This is consistent with common modeling intuition that more data can lead to both better calibrated process models and better trained ML models \cite{read2019process,jia2019physics}. We also saw the PB0 meta-feature, GLM stratification percentage, as the 2nd most important feature for PB-MTL, and included with less importance in PGDL-MTL. Top sources had lower mean observation temperatures, which possibly indicates either more balanced measurements between surface and deeper depths, or a better spread of observations across seasons, in a given lake. For example, source lakes that use mostly surface temperatures would have higher mean observation temp, and source lakes that have many deeper measurements would have lower mean observation temp.

Inspecting the characteristics of the most frequently selected source lakes could guide future monitoring and MTL modeling efforts. Only eleven unique source models were used to predict almost one third of target lake water temperatures using both PGDL-MTL and PB-MTL, and top source lakes were generally deeper, larger, and more \change{well observed}{well-monitored} compared to rarely or never selected source models (Figure \ref{fig:source_property_figure}). Differences between source and target in lake depth and area, as well as the observation count of source lakes, were important meta-features used to select source lakes. While these features likely explain why some lake models are generally more transferable, the unique properties of some target lakes and their selected source models are important to consider when designing lake monitoring campaigns or evaluating future model transfer methods. For example, PGDL source models that were rarely selected (chosen one, two, or three times as a top source) still helped overall test lake performance and were often better actually-ranked options for their target lakes compared to the ranks of commonly chosen (ten or more times) source models (based on ranking the performance of all possible source model transfers to each target lake; median actual PGDL-MTL rank for rare source transfers: 20 of 145, and common source transfers: 36.5 of 145; n=100 and 103, respectively). This pattern did not hold for PB-MTL transfers, with generally worse actual ranks for rare sources compared to common sources (median rank for rare and common were 28.5 and 23, and n=80 and 95, respectively), and additional research may be necessary to understand these differences. The important meta-features used in this study (e.g., differences in maximum depth and area, and the number of observations used to train or calibrate the source model) differ from previous process-based modeling parameter transfer methods that have been applied to rivers. These previous works have instead focused on spatial proximity, spatial fields of hydrologic signatures, or global parameterization \cite{mizukami2017towards}.


Because lake temperature is an ecological ``master factor" \cite{magnuson1979temperature}, predictions at broader scales can support a wide variety of science and management efforts, from improved modeling of biota \cite{jones2006forecasting, hansen2017projected} to improved thermoelectric power plant heat management \cite{cook2015assessing}. PGDL-MTL models can output predictions at scale wherever meteorological and essential lake attribute data are available, and the MTL approach could eventually be developed into use for forecasting applications. A forecasting variant of MTL could be developed by building base models specifically for forecasting (e.g. with probabilistic outputs), and optimizing transfer performance to new systems by simulating forecasting performance instead of hindcasting RMSE. Below, we discuss the various ways the MTL approach can scale to other systems.

The applicability of MTL scales beyond just \change{unobserved}{unmonitored} systems to a large range of \change{observed}{monitored} systems as well, bridging the gap between local accuracy and broad-scale modeling. In Experiment 2, we investigated the point at which, for sparsely \change{observed}{monitored} systems, it would be better to transfer models from different \change{better-observed}{better-monitored} systems as opposed to training PGDL models on what little target data is available. This is a pertinent question for broad scale modeling; while a majority of lakes in this region are \change{unobserved}{unmonitored}, a large fraction of \change{observed}{monitored} lakes have \textless40 observations \cite{stanley2019biases}. Though PGDL models have been shown to outperform calibrated process-based models on even a small number of water temperature sampling dates by taking advantage of process-based simulation data and process-informed learning constraints \cite{read2019process}, MTL presents the opportunity to improve prediction by harnessing  more simulation data, observation data, and metadata from past modeling experiences across many other lake systems. There is also opportunity to expand the MTL framework to incorporate sparse data available in many lakes, where the transferred source models could be fine-tuned using data from the target lake itself.

\begin{table}[h!]
    \centering
    \caption{\textit{Method Comparison Across Broad-Scale Modeling of 1882 Lakes in the Midwestern United States}}
    \scalebox{0.8}{
    \begin{tabular}{|c|c|c|c|}
        \toprule
         Method & \thead{Median RMSE\\ (C)} & \thead{Lower\\Quartile RMSE} & \thead{Upper\\Quartile RMSE} \\
         \midrule
         PB0 & 2.28 & 1.84 & 2.94 \\ 
         \midrule 
         PGDL-MTL & 2.06 & 1.59 & 2.74\\
         \midrule 
         PGDL-MTL9 & 1.80 & 1.40 & 2.38\\
         \bottomrule
    \end{tabular}}

    \label{tab:broad-scale-result}
\end{table}
\begin{figure}[h!]
    \centering
    \includegraphics[width=\textwidth]{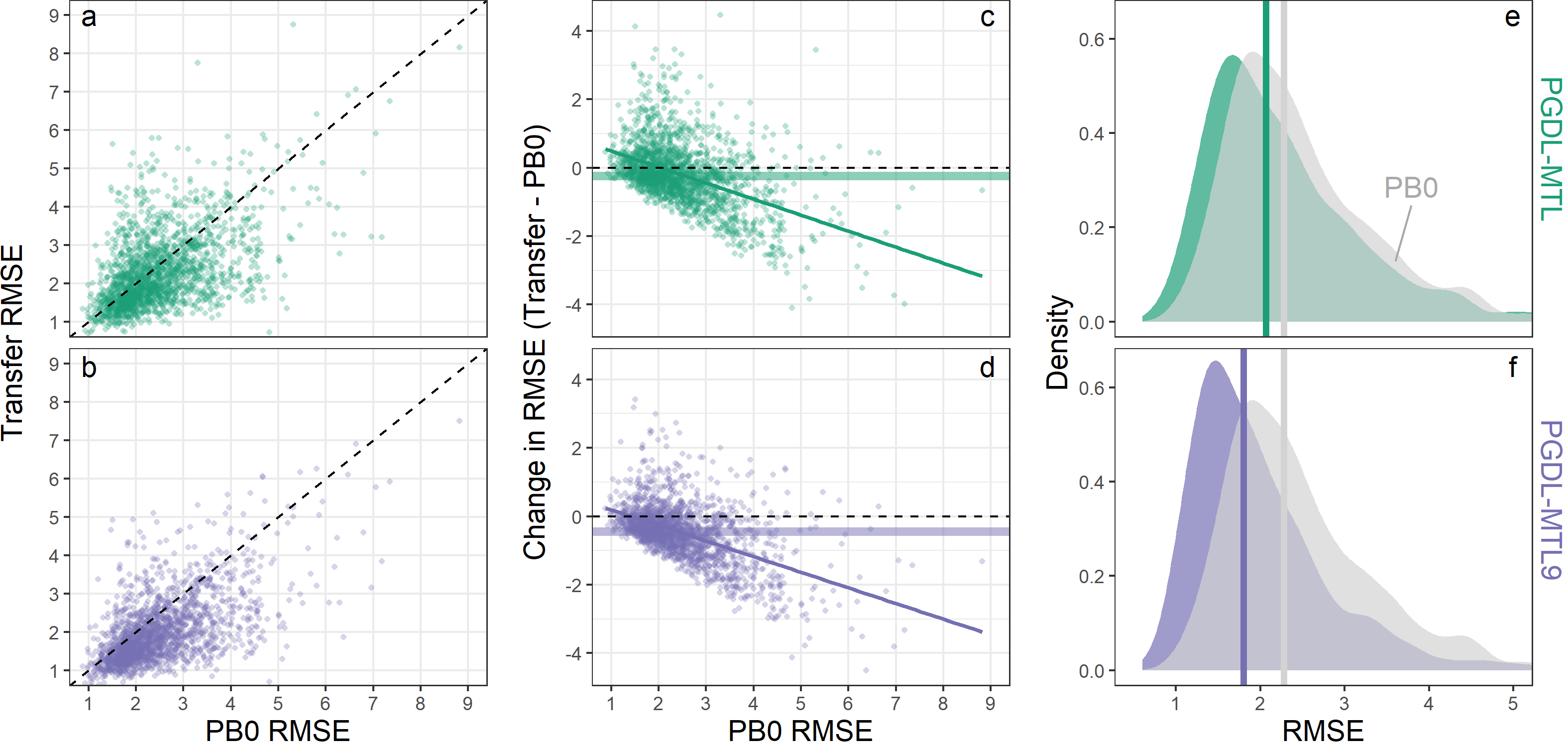}
    \caption{Comparison of model performance \change{of PB0, PGDL-MTL, and PGDL-MTL9 RMSE across 1882 test lakes}{of PGDL-MTL (a,c,e) and PGDL-MTL9 (b,d,f) RMSE relative to PB0 across 1882 test lakes}. a-b) RMSE of PB0 relative to the two transfer models, where the dotted line shows the 1:1 relationship. c-d) The difference between RMSE of the transfer and PB0 models, where the black dotted line shows the zero or no change line, and the solid colored lines show the linear regression fit of the change in RMSE as a function of PB0 RMSE. e-f) The distribution of RMSE from PB0 and transfer models, where the vertical gray and colored lines are the median PB0 and transfer RMSE, respectively.}
    \label{fig:expanded_multipanel}
\end{figure}

Another major benefit of MTL with PGDL in particular is the scalability and efficiency of ML models once the meta-learning model and source models are trained. MTL can be built with data that are easier to obtain than temperature observations (e.g. maximum depth and surface area), and MTL does not require any new models to be trained. Therefore, it can scale to a much larger number of lakes than the ones used in this study. To demonstrate this scalability, we applied the transfer models to 1882 additional lakes that were less \change{observed}{monitored} than our initial 305 lakes. The transfers maintained a significant accuracy improvement over a purely process-based modeling approach (PB0). For this expanded set of lakes, median RMSE was 1.80 \textdegree C for PGDL-MTL9, 2.06 \textdegree C for PGDL-MTL, and 2.29 \textdegree C for PB0 (Table \ref{tab:broad-scale-result}). Temperatures in a majority of lakes were more accurately predicted by the transfer models compared to PB0; for PGDL-MTL9 1484 of 1882 lakes improved over PB0, and for PGDL-MTL 1206 of 1882 improved over PB0 (Figure \ref{fig:expanded_multipanel}). 


Finally, given the demonstrated generalizability of PGDL and PB models using MTL, this approach opens doors to new research directions, like transferring source models into new spatial domains, \add{including remote sensing surface observation data}, incorporating uncertainty quantification, and aggregating models more effectively. Though our study was limited to 5 Midwestern states in the United States, this could be expanded to include a much larger variety of lake types and locations. A remaining question for this transfer approach is, when expanding to new types of lakes, how should an optimal set of source lakes be identified? Another research direction includes uncertainty estimation in the  metamodel construction. Uncertainty estimates could be used to reject a target lake for which all the source model error estimates are confidently high. Furthermore, the ensembling approach could be improved, using more complex methods than a simple average to combine top source models. One promising option is generalized stacking of neural networks \cite{ghorbani2001stacked}, where all the source neural networks would be connected by an averaging layer. \add{Remote sensing data integration could also help in adding surface temperature data to the source models and could allow corrective measures to be taken for predictions in lakes unmonitored by in-situ data. Though, remote sensing observations have known drawbacks in this application such as being limited to only surface temperature on larger lakes \protect\cite{schaeffer2018initial,topp2020research}}.


Given the successful prediction of environmental variables using MTL approaches, there are many research opportunities in different types of applications and data scenarios. For example, predicting only lake surface temperature would allow for the use of MTL without the need for maximum depth measurements, which could allow for predictions in many more lakes. Also, different types of source models could also be used in different scenarios. Some process-based models likely work better for some lakes than others; for example, process models built specifically for reservoir dynamics could be important source models in regions where reservoirs are a common lake type. Other environmental variables could also be targeted for prediction like water quality (e.g. dissolved oxygen, conductivity) and water quantity in lakes, streams, wetlands and other water bodies.

\section*{Data and Code Availability Statement}
The experimental data from this manuscript are freely available in the USGS data release ``Predicting Water Temperature Dynamics of Unmonitored Lakes with Meta Transfer Learning" (\url{https://doi.org/10.5066/P9I00WFR}) \cite{willard2020data}. Also the code used in this is study is available at \url{https://github.com/jdwillard19/MTL_lakes}.

\acknowledgments
 This work is supported by NSF grant \#1934721 under the Harnessing the Data Revolution (HDR) program. The authors acknowledge the Minnesota Supercomputing Institute (MSI) at the University of Minnesota for providing resources that contributed to the research results reported within this paper. URL: \url{http://www.msi.umn.edu}. We thank Jennifer Fair and William Farmer for initial review\add{, and Hayley Corson-Dosch for visualization of Figure \protect\ref{fig:map}}. We also thank the Department of the Interior North Central Climate Adaptation Science Center for funding and the USGS Advanced Research Computing, USGS Yeti Supercomputer (\url{https://doi.org/10.5066/F7D798MJ}) for infrastructure used for GLM simulations. Any use of trade, firm, or product names is for descriptive purposes only and does not imply endorsement by the U.S. Government.

\nocite{patro2015normalization,
        karpatne2017physics,
        karpatne2017theory,
        glorot2010understanding}

\bibliography{agu_journal_main_text}

\begin{thebibliography}{}

\bibitem [\protect \citeauthoryear {%
Aguilera%
\ \protect \BOthers {.}}{%
Aguilera%
\ \protect \BOthers {.}}{%
{\protect \APACyear {2016}}%
}]{%
aguilera2016using}
\APACinsertmetastar {%
aguilera2016using}%
\begin{APACrefauthors}%
Aguilera, R.%
, Livingstone, D\BPBI M.%
, Marc{\'e}, R.%
, Jennings, E.%
, Piera, J.%
\BCBL {}\ \BBA {} Adrian, R.%
\end{APACrefauthors}%
\unskip\
\newblock
\APACrefYearMonthDay{2016}{}{}.
\newblock
{\BBOQ}\APACrefatitle {Using dynamic factor analysis to show how sampling
  resolution and data gaps affect the recognition of patterns in limnological
  time series} {Using dynamic factor analysis to show how sampling resolution
  and data gaps affect the recognition of patterns in limnological time
  series}.{\BBCQ}
\newblock
\APACjournalVolNumPages{Inland Waters}{6}{3}{284--294}.
\PrintBackRefs{\CurrentBib}

\bibitem [\protect \citeauthoryear {%
Archfield%
\ \protect \BOthers {.}}{%
Archfield%
\ \protect \BOthers {.}}{%
{\protect \APACyear {2015}}%
}]{%
archfield2015accelerating}
\APACinsertmetastar {%
archfield2015accelerating}%
\begin{APACrefauthors}%
Archfield, S\BPBI A.%
, Clark, M.%
, Arheimer, B.%
, Hay, L\BPBI E.%
, McMillan, H.%
, Kiang, J\BPBI E.%
\BDBL {}others%
\end{APACrefauthors}%
\unskip\
\newblock
\APACrefYearMonthDay{2015}{}{}.
\newblock
{\BBOQ}\APACrefatitle {Accelerating advances in continental domain hydrologic
  modeling} {Accelerating advances in continental domain hydrologic
  modeling}.{\BBCQ}
\newblock
\APACjournalVolNumPages{Water Resources Research}{51}{12}{10078--10091}.
\PrintBackRefs{\CurrentBib}

\bibitem [\protect \citeauthoryear {%
Baines%
, Webster%
, Kratz%
, Carpenter%
\BCBL {}\ \BBA {} Magnuson%
}{%
Baines%
\ \protect \BOthers {.}}{%
{\protect \APACyear {2000}}%
}]{%
baines2000synchronous}
\APACinsertmetastar {%
baines2000synchronous}%
\begin{APACrefauthors}%
Baines, S\BPBI B.%
, Webster, K\BPBI E.%
, Kratz, T\BPBI K.%
, Carpenter, S\BPBI R.%
\BCBL {}\ \BBA {} Magnuson, J\BPBI J.%
\end{APACrefauthors}%
\unskip\
\newblock
\APACrefYearMonthDay{2000}{}{}.
\newblock
{\BBOQ}\APACrefatitle {Synchronous behavior of temperature, calcium, and
  chlorophyll in lakes of northern Wisconsin} {Synchronous behavior of
  temperature, calcium, and chlorophyll in lakes of northern wisconsin}.{\BBCQ}
\newblock
\APACjournalVolNumPages{Ecology}{81}{3}{815--825}.
\PrintBackRefs{\CurrentBib}

\bibitem [\protect \citeauthoryear {%
Benson%
\ \protect \BOthers {.}}{%
Benson%
\ \protect \BOthers {.}}{%
{\protect \APACyear {2000}}%
}]{%
benson2000regional}
\APACinsertmetastar {%
benson2000regional}%
\begin{APACrefauthors}%
Benson, B\BPBI J.%
, Lenters\&, J\BPBI D.%
, dagger%
, Magnuson, J\BPBI J.%
, Stubbs, M.%
, Dagger%
\BDBL {}Lathrop, R\BPBI C.%
\end{APACrefauthors}%
\unskip\
\newblock
\APACrefYearMonthDay{2000}{}{}.
\newblock
{\BBOQ}\APACrefatitle {Regional coherence of climatic and lake thermal
  variables of four lake districts in the Upper Great Lakes Region of North
  America} {Regional coherence of climatic and lake thermal variables of four
  lake districts in the upper great lakes region of north america}.{\BBCQ}
\newblock
\APACjournalVolNumPages{Freshwater Biology}{43}{3}{517--527}.
\PrintBackRefs{\CurrentBib}

\bibitem [\protect \citeauthoryear {%
Buitinck%
\ \protect \BOthers {.}}{%
Buitinck%
\ \protect \BOthers {.}}{%
{\protect \APACyear {2013}}%
}]{%
buitinck2013api}
\APACinsertmetastar {%
buitinck2013api}%
\begin{APACrefauthors}%
Buitinck, L.%
, Louppe, G.%
, Blondel, M.%
, Pedregosa, F.%
, Mueller, A.%
, Grisel, O.%
\BDBL {}others%
\end{APACrefauthors}%
\unskip\
\newblock
\APACrefYearMonthDay{2013}{}{}.
\newblock
{\BBOQ}\APACrefatitle {API design for machine learning software: experiences
  from the scikit-learn project} {Api design for machine learning software:
  experiences from the scikit-learn project}.{\BBCQ}
\newblock
\APACjournalVolNumPages{arXiv preprint arXiv:1309.0238}{}{}{}.
\PrintBackRefs{\CurrentBib}

\bibitem [\protect \citeauthoryear {%
Castiello%
, Castellano%
\BCBL {}\ \BBA {} Fanelli%
}{%
Castiello%
\ \protect \BOthers {.}}{%
{\protect \APACyear {2005}}%
}]{%
castiello2005meta}
\APACinsertmetastar {%
castiello2005meta}%
\begin{APACrefauthors}%
Castiello, C.%
, Castellano, G.%
\BCBL {}\ \BBA {} Fanelli, A\BPBI M.%
\end{APACrefauthors}%
\unskip\
\newblock
\APACrefYearMonthDay{2005}{}{}.
\newblock
{\BBOQ}\APACrefatitle {Meta-data: Characterization of input features for
  meta-learning} {Meta-data: Characterization of input features for
  meta-learning}.{\BBCQ}
\newblock
\BIn{} \APACrefbtitle {International Conference on Modeling Decisions for
  Artificial Intelligence} {International conference on modeling decisions for
  artificial intelligence}\ (\BPGS\ 457--468).
\PrintBackRefs{\CurrentBib}

\bibitem [\protect \citeauthoryear {%
Caughlan%
\ \BBA {} Oakley%
}{%
Caughlan%
\ \BBA {} Oakley%
}{%
{\protect \APACyear {2001}}%
}]{%
caughlan2001cost}
\APACinsertmetastar {%
caughlan2001cost}%
\begin{APACrefauthors}%
Caughlan, L.%
\BCBT {}\ \BBA {} Oakley, K\BPBI L.%
\end{APACrefauthors}%
\unskip\
\newblock
\APACrefYearMonthDay{2001}{}{}.
\newblock
{\BBOQ}\APACrefatitle {Cost considerations for long-term ecological monitoring}
  {Cost considerations for long-term ecological monitoring}.{\BBCQ}
\newblock
\APACjournalVolNumPages{Ecological indicators}{1}{2}{123--134}.
\PrintBackRefs{\CurrentBib}

\bibitem [\protect \citeauthoryear {%
Cook%
, King%
, Davidson%
\BCBL {}\ \BBA {} Webber%
}{%
Cook%
\ \protect \BOthers {.}}{%
{\protect \APACyear {2015}}%
}]{%
cook2015assessing}
\APACinsertmetastar {%
cook2015assessing}%
\begin{APACrefauthors}%
Cook, M\BPBI A.%
, King, C\BPBI W.%
, Davidson, F\BPBI T.%
\BCBL {}\ \BBA {} Webber, M\BPBI E.%
\end{APACrefauthors}%
\unskip\
\newblock
\APACrefYearMonthDay{2015}{}{}.
\newblock
{\BBOQ}\APACrefatitle {Assessing the impacts of droughts and heat waves at
  thermoelectric power plants in the United States using integrated regression,
  thermodynamic, and climate models} {Assessing the impacts of droughts and
  heat waves at thermoelectric power plants in the united states using
  integrated regression, thermodynamic, and climate models}.{\BBCQ}
\newblock
\APACjournalVolNumPages{Energy Reports}{1}{}{193--203}.
\PrintBackRefs{\CurrentBib}

\bibitem [\protect \citeauthoryear {%
Cuddington%
\ \protect \BOthers {.}}{%
Cuddington%
\ \protect \BOthers {.}}{%
{\protect \APACyear {2013}}%
}]{%
cuddington2013process}
\APACinsertmetastar {%
cuddington2013process}%
\begin{APACrefauthors}%
Cuddington, K.%
, Fortin, M\BHBI J.%
, Gerber, L.%
, Hastings, A.%
, Liebhold, A.%
, O'connor, M.%
\BCBL {}\ \BBA {} Ray, C.%
\end{APACrefauthors}%
\unskip\
\newblock
\APACrefYearMonthDay{2013}{}{}.
\newblock
{\BBOQ}\APACrefatitle {Process-based models are required to manage ecological
  systems in a changing world} {Process-based models are required to manage
  ecological systems in a changing world}.{\BBCQ}
\newblock
\APACjournalVolNumPages{Ecosphere}{4}{2}{1--12}.
\PrintBackRefs{\CurrentBib}

\bibitem [\protect \citeauthoryear {%
Dugdale%
, Hannah%
\BCBL {}\ \BBA {} Malcolm%
}{%
Dugdale%
\ \protect \BOthers {.}}{%
{\protect \APACyear {2017}}%
}]{%
dugdale2017river}
\APACinsertmetastar {%
dugdale2017river}%
\begin{APACrefauthors}%
Dugdale, S\BPBI J.%
, Hannah, D\BPBI M.%
\BCBL {}\ \BBA {} Malcolm, I\BPBI A.%
\end{APACrefauthors}%
\unskip\
\newblock
\APACrefYearMonthDay{2017}{}{}.
\newblock
{\BBOQ}\APACrefatitle {River temperature modelling: A review of process-based
  approaches and future directions} {River temperature modelling: A review of
  process-based approaches and future directions}.{\BBCQ}
\newblock
\APACjournalVolNumPages{Earth-Science Reviews}{175}{}{97--113}.
\PrintBackRefs{\CurrentBib}

\bibitem [\protect \citeauthoryear {%
Erdal%
\ \BBA {} Karakurt%
}{%
Erdal%
\ \BBA {} Karakurt%
}{%
{\protect \APACyear {2013}}%
}]{%
erdal2013advancing}
\APACinsertmetastar {%
erdal2013advancing}%
\begin{APACrefauthors}%
Erdal, H\BPBI I.%
\BCBT {}\ \BBA {} Karakurt, O.%
\end{APACrefauthors}%
\unskip\
\newblock
\APACrefYearMonthDay{2013}{}{}.
\newblock
{\BBOQ}\APACrefatitle {Advancing monthly streamflow prediction accuracy of CART
  models using ensemble learning paradigms} {Advancing monthly streamflow
  prediction accuracy of cart models using ensemble learning paradigms}.{\BBCQ}
\newblock
\APACjournalVolNumPages{Journal of Hydrology}{477}{}{119--128}.
\PrintBackRefs{\CurrentBib}

\bibitem [\protect \citeauthoryear {%
Erlandsson%
\ \protect \BOthers {.}}{%
Erlandsson%
\ \protect \BOthers {.}}{%
{\protect \APACyear {2008}}%
}]{%
erlandsson2008thirty}
\APACinsertmetastar {%
erlandsson2008thirty}%
\begin{APACrefauthors}%
Erlandsson, M.%
, Buffam, I.%
, F{\"o}lster, J.%
, Laudon, H.%
, Temnerud, J.%
, Weyhenmeyer, G\BPBI A.%
\BCBL {}\ \BBA {} Bishop, K.%
\end{APACrefauthors}%
\unskip\
\newblock
\APACrefYearMonthDay{2008}{}{}.
\newblock
{\BBOQ}\APACrefatitle {Thirty-five years of synchrony in the organic matter
  concentrations of Swedish rivers explained by variation in flow and sulphate}
  {Thirty-five years of synchrony in the organic matter concentrations of
  swedish rivers explained by variation in flow and sulphate}.{\BBCQ}
\newblock
\APACjournalVolNumPages{Global Change Biology}{14}{5}{1191--1198}.
\PrintBackRefs{\CurrentBib}

\bibitem [\protect \citeauthoryear {%
Fatichi%
\ \protect \BOthers {.}}{%
Fatichi%
\ \protect \BOthers {.}}{%
{\protect \APACyear {2016}}%
}]{%
fatichi2016overview}
\APACinsertmetastar {%
fatichi2016overview}%
\begin{APACrefauthors}%
Fatichi, S.%
, Vivoni, E\BPBI R.%
, Ogden, F\BPBI L.%
, Ivanov, V\BPBI Y.%
, Mirus, B.%
, Gochis, D.%
\BDBL {}others%
\end{APACrefauthors}%
\unskip\
\newblock
\APACrefYearMonthDay{2016}{}{}.
\newblock
{\BBOQ}\APACrefatitle {An overview of current applications, challenges, and
  future trends in distributed process-based models in hydrology} {An overview
  of current applications, challenges, and future trends in distributed
  process-based models in hydrology}.{\BBCQ}
\newblock
\APACjournalVolNumPages{Journal of Hydrology}{537}{}{45--60}.
\PrintBackRefs{\CurrentBib}

\bibitem [\protect \citeauthoryear {%
Fink%
, Schmid%
, Wahl%
, Wolf%
\BCBL {}\ \BBA {} W{\"u}est%
}{%
Fink%
\ \protect \BOthers {.}}{%
{\protect \APACyear {2014}}%
}]{%
fink2014heat}
\APACinsertmetastar {%
fink2014heat}%
\begin{APACrefauthors}%
Fink, G.%
, Schmid, M.%
, Wahl, B.%
, Wolf, T.%
\BCBL {}\ \BBA {} W{\"u}est, A.%
\end{APACrefauthors}%
\unskip\
\newblock
\APACrefYearMonthDay{2014}{}{}.
\newblock
{\BBOQ}\APACrefatitle {Heat flux modifications related to climate-induced
  warming of large European lakes} {Heat flux modifications related to
  climate-induced warming of large european lakes}.{\BBCQ}
\newblock
\APACjournalVolNumPages{Water Resources Research}{50}{3}{2072--2085}.
\PrintBackRefs{\CurrentBib}

\bibitem [\protect \citeauthoryear {%
Friedman%
}{%
Friedman%
}{%
{\protect \APACyear {2001}}%
}]{%
friedman2001greedy}
\APACinsertmetastar {%
friedman2001greedy}%
\begin{APACrefauthors}%
Friedman, J\BPBI H.%
\end{APACrefauthors}%
\unskip\
\newblock
\APACrefYearMonthDay{2001}{}{}.
\newblock
{\BBOQ}\APACrefatitle {Greedy function approximation: a gradient boosting
  machine} {Greedy function approximation: a gradient boosting machine}.{\BBCQ}
\newblock
\APACjournalVolNumPages{Annals of statistics}{}{}{1189--1232}.
\PrintBackRefs{\CurrentBib}

\bibitem [\protect \citeauthoryear {%
Gaudard%
, R{\aa}man~Vinn{\aa}%
, B{\"a}renbold%
, Schmid%
\BCBL {}\ \BBA {} Bouffard%
}{%
Gaudard%
\ \protect \BOthers {.}}{%
{\protect \APACyear {2019}}%
}]{%
gaudard2019toward}
\APACinsertmetastar {%
gaudard2019toward}%
\begin{APACrefauthors}%
Gaudard, A.%
, R{\aa}man~Vinn{\aa}, L.%
, B{\"a}renbold, F.%
, Schmid, M.%
\BCBL {}\ \BBA {} Bouffard, D.%
\end{APACrefauthors}%
\unskip\
\newblock
\APACrefYearMonthDay{2019}{}{}.
\newblock
{\BBOQ}\APACrefatitle {Toward an open access to high-frequency lake modeling
  and statistics data for scientists and practitioners--the case of Swiss lakes
  using Simstrat v2. 1.} {Toward an open access to high-frequency lake modeling
  and statistics data for scientists and practitioners--the case of swiss lakes
  using simstrat v2. 1.}{\BBCQ}
\newblock
\APACjournalVolNumPages{Geoscientific Model Development}{12}{9}{}.
\PrintBackRefs{\CurrentBib}

\bibitem [\protect \citeauthoryear {%
George%
, Talling%
\BCBL {}\ \BBA {} Rigg%
}{%
George%
\ \protect \BOthers {.}}{%
{\protect \APACyear {2000}}%
}]{%
george2000factors}
\APACinsertmetastar {%
george2000factors}%
\begin{APACrefauthors}%
George, D.%
, Talling, J.%
\BCBL {}\ \BBA {} Rigg, E.%
\end{APACrefauthors}%
\unskip\
\newblock
\APACrefYearMonthDay{2000}{}{}.
\newblock
{\BBOQ}\APACrefatitle {Factors influencing the temporal coherence of five lakes
  in the English Lake District} {Factors influencing the temporal coherence of
  five lakes in the english lake district}.{\BBCQ}
\newblock
\APACjournalVolNumPages{Freshwater Biology}{43}{3}{449--461}.
\PrintBackRefs{\CurrentBib}

\bibitem [\protect \citeauthoryear {%
Ghorbani%
\ \BBA {} Owrangh%
}{%
Ghorbani%
\ \BBA {} Owrangh%
}{%
{\protect \APACyear {2001}}%
}]{%
ghorbani2001stacked}
\APACinsertmetastar {%
ghorbani2001stacked}%
\begin{APACrefauthors}%
Ghorbani, A\BPBI A.%
\BCBT {}\ \BBA {} Owrangh, K.%
\end{APACrefauthors}%
\unskip\
\newblock
\APACrefYearMonthDay{2001}{}{}.
\newblock
{\BBOQ}\APACrefatitle {Stacked generalization in neural networks:
  generalization on statistically neutral problems} {Stacked generalization in
  neural networks: generalization on statistically neutral problems}.{\BBCQ}
\newblock
\BIn{} \APACrefbtitle {IJCNN'01. International Joint Conference on Neural
  Networks. Proceedings (Cat. No. 01CH37222)} {Ijcnn'01. international joint
  conference on neural networks. proceedings (cat. no. 01ch37222)}\ (\BVOL~3,
  \BPGS\ 1715--1720).
\PrintBackRefs{\CurrentBib}

\bibitem [\protect \citeauthoryear {%
Glorot%
\ \BBA {} Bengio%
}{%
Glorot%
\ \BBA {} Bengio%
}{%
{\protect \APACyear {2010}}%
}]{%
glorot2010understanding}
\APACinsertmetastar {%
glorot2010understanding}%
\begin{APACrefauthors}%
Glorot, X.%
\BCBT {}\ \BBA {} Bengio, Y.%
\end{APACrefauthors}%
\unskip\
\newblock
\APACrefYearMonthDay{2010}{}{}.
\newblock
{\BBOQ}\APACrefatitle {Understanding the difficulty of training deep
  feedforward neural networks} {Understanding the difficulty of training deep
  feedforward neural networks}.{\BBCQ}
\newblock
\BIn{} \APACrefbtitle {Proceedings of the thirteenth international conference
  on artificial intelligence and statistics} {Proceedings of the thirteenth
  international conference on artificial intelligence and statistics}\ (\BPGS\
  249--256).
\PrintBackRefs{\CurrentBib}

\bibitem [\protect \citeauthoryear {%
Goodfellow%
, Bengio%
, Courville%
\BCBL {}\ \BBA {} Bengio%
}{%
Goodfellow%
\ \protect \BOthers {.}}{%
{\protect \APACyear {2016}}%
}]{%
goodfellow2016deep}
\APACinsertmetastar {%
goodfellow2016deep}%
\begin{APACrefauthors}%
Goodfellow, I.%
, Bengio, Y.%
, Courville, A.%
\BCBL {}\ \BBA {} Bengio, Y.%
\end{APACrefauthors}%
\unskip\
\newblock
\APACrefYear{2016}.
\newblock
\APACrefbtitle {Deep learning} {Deep learning}\ (\BVOL~1).
\newblock
\APACaddressPublisher{}{MIT press Cambridge}.
\PrintBackRefs{\CurrentBib}

\bibitem [\protect \citeauthoryear {%
Gorham%
\ \BBA {} Boyce%
}{%
Gorham%
\ \BBA {} Boyce%
}{%
{\protect \APACyear {1989}}%
}]{%
gorham1989influence}
\APACinsertmetastar {%
gorham1989influence}%
\begin{APACrefauthors}%
Gorham, E.%
\BCBT {}\ \BBA {} Boyce, F\BPBI M.%
\end{APACrefauthors}%
\unskip\
\newblock
\APACrefYearMonthDay{1989}{}{}.
\newblock
{\BBOQ}\APACrefatitle {Influence of lake surface area and depth upon thermal
  stratification and the depth of the summer thermocline} {Influence of lake
  surface area and depth upon thermal stratification and the depth of the
  summer thermocline}.{\BBCQ}
\newblock
\APACjournalVolNumPages{Journal of Great Lakes Research}{15}{2}{233--245}.
\PrintBackRefs{\CurrentBib}

\bibitem [\protect \citeauthoryear {%
Guyon%
, Weston%
, Barnhill%
\BCBL {}\ \BBA {} Vapnik%
}{%
Guyon%
\ \protect \BOthers {.}}{%
{\protect \APACyear {2002}}%
}]{%
guyon2002gene}
\APACinsertmetastar {%
guyon2002gene}%
\begin{APACrefauthors}%
Guyon, I.%
, Weston, J.%
, Barnhill, S.%
\BCBL {}\ \BBA {} Vapnik, V.%
\end{APACrefauthors}%
\unskip\
\newblock
\APACrefYearMonthDay{2002}{}{}.
\newblock
{\BBOQ}\APACrefatitle {Gene selection for cancer classification using support
  vector machines} {Gene selection for cancer classification using support
  vector machines}.{\BBCQ}
\newblock
\APACjournalVolNumPages{Machine learning}{46}{1-3}{389--422}.
\PrintBackRefs{\CurrentBib}

\bibitem [\protect \citeauthoryear {%
Hampton%
\ \protect \BOthers {.}}{%
Hampton%
\ \protect \BOthers {.}}{%
{\protect \APACyear {2013}}%
}]{%
hampton2013big}
\APACinsertmetastar {%
hampton2013big}%
\begin{APACrefauthors}%
Hampton, S\BPBI E.%
, Strasser, C\BPBI A.%
, Tewksbury, J\BPBI J.%
, Gram, W\BPBI K.%
, Budden, A\BPBI E.%
, Batcheller, A\BPBI L.%
\BDBL {}Porter, J\BPBI H.%
\end{APACrefauthors}%
\unskip\
\newblock
\APACrefYearMonthDay{2013}{}{}.
\newblock
{\BBOQ}\APACrefatitle {Big data and the future of ecology} {Big data and the
  future of ecology}.{\BBCQ}
\newblock
\APACjournalVolNumPages{Frontiers in Ecology and the
  Environment}{11}{3}{156--162}.
\PrintBackRefs{\CurrentBib}

\bibitem [\protect \citeauthoryear {%
Hansen%
, Read%
, Hansen%
\BCBL {}\ \BBA {} Winslow%
}{%
Hansen%
\ \protect \BOthers {.}}{%
{\protect \APACyear {2017}}%
}]{%
hansen2017projected}
\APACinsertmetastar {%
hansen2017projected}%
\begin{APACrefauthors}%
Hansen, G\BPBI J.%
, Read, J\BPBI S.%
, Hansen, J\BPBI F.%
\BCBL {}\ \BBA {} Winslow, L\BPBI A.%
\end{APACrefauthors}%
\unskip\
\newblock
\APACrefYearMonthDay{2017}{}{}.
\newblock
{\BBOQ}\APACrefatitle {Projected shifts in fish species dominance in Wisconsin
  lakes under climate change} {Projected shifts in fish species dominance in
  wisconsin lakes under climate change}.{\BBCQ}
\newblock
\APACjournalVolNumPages{Global change biology}{23}{4}{1463--1476}.
\PrintBackRefs{\CurrentBib}

\bibitem [\protect \citeauthoryear {%
Hipsey%
\ \protect \BOthers {.}}{%
Hipsey%
\ \protect \BOthers {.}}{%
{\protect \APACyear {2019}}%
}]{%
hipsey2019general}
\APACinsertmetastar {%
hipsey2019general}%
\begin{APACrefauthors}%
Hipsey, M\BPBI R.%
, Bruce, L\BPBI C.%
, Boon, C.%
, Busch, B.%
, Carey, C\BPBI C.%
, Hamilton, D\BPBI P.%
\BDBL {}others%
\end{APACrefauthors}%
\unskip\
\newblock
\APACrefYearMonthDay{2019}{}{}.
\newblock
{\BBOQ}\APACrefatitle {A General Lake Model (GLM 3.0) for linking with
  high-frequency sensor data from the Global Lake Ecological Observatory
  Network (GLEON)} {A general lake model (glm 3.0) for linking with
  high-frequency sensor data from the global lake ecological observatory
  network (gleon)}.{\BBCQ}
\newblock
\APACjournalVolNumPages{Geoscientific Model Development}{}{}{}.
\PrintBackRefs{\CurrentBib}

\bibitem [\protect \citeauthoryear {%
Hipsey%
\ \protect \BOthers {.}}{%
Hipsey%
\ \protect \BOthers {.}}{%
{\protect \APACyear {2015}}%
}]{%
hipsey2015predicting}
\APACinsertmetastar {%
hipsey2015predicting}%
\begin{APACrefauthors}%
Hipsey, M\BPBI R.%
, Hamilton, D\BPBI P.%
, Hanson, P\BPBI C.%
, Carey, C\BPBI C.%
, Coletti, J\BPBI Z.%
, Read, J\BPBI S.%
\BDBL {}Brookes, J\BPBI D.%
\end{APACrefauthors}%
\unskip\
\newblock
\APACrefYearMonthDay{2015}{}{}.
\newblock
{\BBOQ}\APACrefatitle {Predicting the resilience and recovery of aquatic
  systems: A framework for model evolution within environmental observatories}
  {Predicting the resilience and recovery of aquatic systems: A framework for
  model evolution within environmental observatories}.{\BBCQ}
\newblock
\APACjournalVolNumPages{Water Resources Research}{51}{9}{7023--7043}.
\PrintBackRefs{\CurrentBib}

\bibitem [\protect \citeauthoryear {%
Hochreiter%
\ \BBA {} Schmidhuber%
}{%
Hochreiter%
\ \BBA {} Schmidhuber%
}{%
{\protect \APACyear {1997}}%
}]{%
hochreiter1997long}
\APACinsertmetastar {%
hochreiter1997long}%
\begin{APACrefauthors}%
Hochreiter, S.%
\BCBT {}\ \BBA {} Schmidhuber, J.%
\end{APACrefauthors}%
\unskip\
\newblock
\APACrefYearMonthDay{1997}{}{}.
\newblock
{\BBOQ}\APACrefatitle {Long short-term memory} {Long short-term memory}.{\BBCQ}
\newblock
\APACjournalVolNumPages{Neural computation}{9}{8}{1735--1780}.
\PrintBackRefs{\CurrentBib}

\bibitem [\protect \citeauthoryear {%
Jia%
\ \protect \BOthers {.}}{%
Jia%
\ \protect \BOthers {.}}{%
{\protect \APACyear {2019}}%
}]{%
jia2019physics}
\APACinsertmetastar {%
jia2019physics}%
\begin{APACrefauthors}%
Jia, X.%
, Willard, J\BPBI D.%
, Karpatne, A.%
, Read, J.%
, Zwart, J.%
, Steinbach, M.%
\BCBL {}\ \BBA {} Kumar, V.%
\end{APACrefauthors}%
\unskip\
\newblock
\APACrefYearMonthDay{2019}{}{}.
\newblock
{\BBOQ}\APACrefatitle {Physics guided RNNs for modeling dynamical systems: A
  case study in simulating lake temperature profiles} {Physics guided rnns for
  modeling dynamical systems: A case study in simulating lake temperature
  profiles}.{\BBCQ}
\newblock
\BIn{} \APACrefbtitle {Proceedings of the 2019 SIAM International Conference on
  Data Mining} {Proceedings of the 2019 siam international conference on data
  mining}\ (\BPGS\ 558--566).
\PrintBackRefs{\CurrentBib}

\bibitem [\protect \citeauthoryear {%
Jones%
, Shuter%
, Zhao%
\BCBL {}\ \BBA {} Stockwell%
}{%
Jones%
\ \protect \BOthers {.}}{%
{\protect \APACyear {2006}}%
}]{%
jones2006forecasting}
\APACinsertmetastar {%
jones2006forecasting}%
\begin{APACrefauthors}%
Jones, M\BPBI L.%
, Shuter, B\BPBI J.%
, Zhao, Y.%
\BCBL {}\ \BBA {} Stockwell, J\BPBI D.%
\end{APACrefauthors}%
\unskip\
\newblock
\APACrefYearMonthDay{2006}{}{}.
\newblock
{\BBOQ}\APACrefatitle {Forecasting effects of climate change on Great Lakes
  fisheries: models that link habitat supply to population dynamics can help}
  {Forecasting effects of climate change on great lakes fisheries: models that
  link habitat supply to population dynamics can help}.{\BBCQ}
\newblock
\APACjournalVolNumPages{Canadian Journal of Fisheries and Aquatic
  Sciences}{63}{2}{457--468}.
\PrintBackRefs{\CurrentBib}

\bibitem [\protect \citeauthoryear {%
Karpatne%
, Atluri%
\BCBL {}\ \protect \BOthers {.}}{%
Karpatne%
, Atluri%
\BCBL {}\ \protect \BOthers {.}}{%
{\protect \APACyear {2017}}%
}]{%
karpatne2017theory}
\APACinsertmetastar {%
karpatne2017theory}%
\begin{APACrefauthors}%
Karpatne, A.%
, Atluri, G.%
, Faghmous, J\BPBI H.%
, Steinbach, M.%
, Banerjee, A.%
, Ganguly, A.%
\BDBL {}Kumar, V.%
\end{APACrefauthors}%
\unskip\
\newblock
\APACrefYearMonthDay{2017}{}{}.
\newblock
{\BBOQ}\APACrefatitle {Theory-guided data science: A new paradigm for
  scientific discovery from data} {Theory-guided data science: A new paradigm
  for scientific discovery from data}.{\BBCQ}
\newblock
\APACjournalVolNumPages{IEEE Transactions on Knowledge and Data
  Engineering}{29}{10}{2318--2331}.
\PrintBackRefs{\CurrentBib}

\bibitem [\protect \citeauthoryear {%
Karpatne%
, Ebert-Uphoff%
, Ravela%
, Babaie%
\BCBL {}\ \BBA {} Kumar%
}{%
Karpatne%
\ \protect \BOthers {.}}{%
{\protect \APACyear {2018}}%
}]{%
karpatne2018machine}
\APACinsertmetastar {%
karpatne2018machine}%
\begin{APACrefauthors}%
Karpatne, A.%
, Ebert-Uphoff, I.%
, Ravela, S.%
, Babaie, H\BPBI A.%
\BCBL {}\ \BBA {} Kumar, V.%
\end{APACrefauthors}%
\unskip\
\newblock
\APACrefYearMonthDay{2018}{}{}.
\newblock
{\BBOQ}\APACrefatitle {Machine learning for the geosciences: Challenges and
  opportunities} {Machine learning for the geosciences: Challenges and
  opportunities}.{\BBCQ}
\newblock
\APACjournalVolNumPages{IEEE Transactions on Knowledge and Data
  Engineering}{}{}{}.
\PrintBackRefs{\CurrentBib}

\bibitem [\protect \citeauthoryear {%
Karpatne%
, Watkins%
, Read%
\BCBL {}\ \BBA {} Kumar%
}{%
Karpatne%
, Watkins%
\BCBL {}\ \protect \BOthers {.}}{%
{\protect \APACyear {2017}}%
}]{%
karpatne2017physics}
\APACinsertmetastar {%
karpatne2017physics}%
\begin{APACrefauthors}%
Karpatne, A.%
, Watkins, W.%
, Read, J.%
\BCBL {}\ \BBA {} Kumar, V.%
\end{APACrefauthors}%
\unskip\
\newblock
\APACrefYearMonthDay{2017}{}{}.
\newblock
{\BBOQ}\APACrefatitle {Physics-guided neural networks (pgnn): An application in
  lake temperature modeling} {Physics-guided neural networks (pgnn): An
  application in lake temperature modeling}.{\BBCQ}
\newblock
\APACjournalVolNumPages{arXiv preprint arXiv:1710.11431}{}{}{}.
\PrintBackRefs{\CurrentBib}

\bibitem [\protect \citeauthoryear {%
Kashinath%
\ \protect \BOthers {.}}{%
Kashinath%
\ \protect \BOthers {.}}{%
{\protect \APACyear {2021}}%
}]{%
kashinath2021physics}
\APACinsertmetastar {%
kashinath2021physics}%
\begin{APACrefauthors}%
Kashinath, K.%
, Mustafa, M.%
, Albert, A.%
, Wu, J.%
, Jiang, C.%
, Esmaeilzadeh, S.%
\BDBL {}others%
\end{APACrefauthors}%
\unskip\
\newblock
\APACrefYearMonthDay{2021}{}{}.
\newblock
{\BBOQ}\APACrefatitle {Physics-informed machine learning: case studies for
  weather and climate modelling} {Physics-informed machine learning: case
  studies for weather and climate modelling}.{\BBCQ}
\newblock
\APACjournalVolNumPages{Philosophical Transactions of the Royal Society
  A}{379}{2194}{20200093}.
\PrintBackRefs{\CurrentBib}

\bibitem [\protect \citeauthoryear {%
Kaya%
\ \protect \BOthers {.}}{%
Kaya%
\ \protect \BOthers {.}}{%
{\protect \APACyear {2019}}%
}]{%
kaya2019analysis}
\APACinsertmetastar {%
kaya2019analysis}%
\begin{APACrefauthors}%
Kaya, A.%
, Keceli, A\BPBI S.%
, Catal, C.%
, Yalic, H\BPBI Y.%
, Temucin, H.%
\BCBL {}\ \BBA {} Tekinerdogan, B.%
\end{APACrefauthors}%
\unskip\
\newblock
\APACrefYearMonthDay{2019}{}{}.
\newblock
{\BBOQ}\APACrefatitle {Analysis of transfer learning for deep neural network
  based plant classification models} {Analysis of transfer learning for deep
  neural network based plant classification models}.{\BBCQ}
\newblock
\APACjournalVolNumPages{Computers and electronics in
  agriculture}{158}{}{20--29}.
\PrintBackRefs{\CurrentBib}

\bibitem [\protect \citeauthoryear {%
Krogh%
\ \BBA {} Vedelsby%
}{%
Krogh%
\ \BBA {} Vedelsby%
}{%
{\protect \APACyear {1995}}%
}]{%
krogh1995neural}
\APACinsertmetastar {%
krogh1995neural}%
\begin{APACrefauthors}%
Krogh, A.%
\BCBT {}\ \BBA {} Vedelsby, J.%
\end{APACrefauthors}%
\unskip\
\newblock
\APACrefYearMonthDay{1995}{}{}.
\newblock
{\BBOQ}\APACrefatitle {Neural network ensembles, cross validation, and active
  learning} {Neural network ensembles, cross validation, and active
  learning}.{\BBCQ}
\newblock
\BIn{} \APACrefbtitle {Advances in neural information processing systems}
  {Advances in neural information processing systems}\ (\BPGS\ 231--238).
\PrintBackRefs{\CurrentBib}

\bibitem [\protect \citeauthoryear {%
Kumar%
, Samaniego%
\BCBL {}\ \BBA {} Attinger%
}{%
Kumar%
\ \protect \BOthers {.}}{%
{\protect \APACyear {2013}}%
}]{%
kumar2013implications}
\APACinsertmetastar {%
kumar2013implications}%
\begin{APACrefauthors}%
Kumar, R.%
, Samaniego, L.%
\BCBL {}\ \BBA {} Attinger, S.%
\end{APACrefauthors}%
\unskip\
\newblock
\APACrefYearMonthDay{2013}{}{}.
\newblock
{\BBOQ}\APACrefatitle {Implications of distributed hydrologic model
  parameterization on water fluxes at multiple scales and locations}
  {Implications of distributed hydrologic model parameterization on water
  fluxes at multiple scales and locations}.{\BBCQ}
\newblock
\APACjournalVolNumPages{Water Resources Research}{49}{1}{360--379}.
\PrintBackRefs{\CurrentBib}

\bibitem [\protect \citeauthoryear {%
Kuncheva%
\ \BBA {} Whitaker%
}{%
Kuncheva%
\ \BBA {} Whitaker%
}{%
{\protect \APACyear {2003}}%
}]{%
kuncheva2003measures}
\APACinsertmetastar {%
kuncheva2003measures}%
\begin{APACrefauthors}%
Kuncheva, L\BPBI I.%
\BCBT {}\ \BBA {} Whitaker, C\BPBI J.%
\end{APACrefauthors}%
\unskip\
\newblock
\APACrefYearMonthDay{2003}{}{}.
\newblock
{\BBOQ}\APACrefatitle {Measures of diversity in classifier ensembles and their
  relationship with the ensemble accuracy} {Measures of diversity in classifier
  ensembles and their relationship with the ensemble accuracy}.{\BBCQ}
\newblock
\APACjournalVolNumPages{Machine learning}{51}{2}{181--207}.
\PrintBackRefs{\CurrentBib}

\bibitem [\protect \citeauthoryear {%
X\BHBI p.~Liu%
, Zhang%
, Lu%
\BCBL {}\ \BBA {} Zhang%
}{%
X\BHBI p.~Liu%
\ \protect \BOthers {.}}{%
{\protect \APACyear {2019}}%
}]{%
liu2019risk}
\APACinsertmetastar {%
liu2019risk}%
\begin{APACrefauthors}%
Liu, X\BHBI p.%
, Zhang, G\BHBI q.%
, Lu, J.%
\BCBL {}\ \BBA {} Zhang, J\BHBI q.%
\end{APACrefauthors}%
\unskip\
\newblock
\APACrefYearMonthDay{2019}{}{}.
\newblock
{\BBOQ}\APACrefatitle {Risk assessment using transfer learning for grassland
  fires} {Risk assessment using transfer learning for grassland fires}.{\BBCQ}
\newblock
\APACjournalVolNumPages{Agricultural and forest meteorology}{269}{}{102--111}.
\PrintBackRefs{\CurrentBib}

\bibitem [\protect \citeauthoryear {%
Y.~Liu%
\ \protect \BOthers {.}}{%
Y.~Liu%
\ \protect \BOthers {.}}{%
{\protect \APACyear {2017}}%
}]{%
liu2017review}
\APACinsertmetastar {%
liu2017review}%
\begin{APACrefauthors}%
Liu, Y.%
, Engel, B\BPBI A.%
, Flanagan, D\BPBI C.%
, Gitau, M\BPBI W.%
, McMillan, S\BPBI K.%
\BCBL {}\ \BBA {} Chaubey, I.%
\end{APACrefauthors}%
\unskip\
\newblock
\APACrefYearMonthDay{2017}{}{}.
\newblock
{\BBOQ}\APACrefatitle {A review on effectiveness of best management practices
  in improving hydrology and water quality: needs and opportunities} {A review
  on effectiveness of best management practices in improving hydrology and
  water quality: needs and opportunities}.{\BBCQ}
\newblock
\APACjournalVolNumPages{Science of the Total Environment}{601}{}{580--593}.
\PrintBackRefs{\CurrentBib}

\bibitem [\protect \citeauthoryear {%
Livingstone%
}{%
Livingstone%
}{%
{\protect \APACyear {2008}}%
}]{%
livingstone2008change}
\APACinsertmetastar {%
livingstone2008change}%
\begin{APACrefauthors}%
Livingstone, D\BPBI M.%
\end{APACrefauthors}%
\unskip\
\newblock
\APACrefYearMonthDay{2008}{}{}.
\newblock
{\BBOQ}\APACrefatitle {A change of climate provokes a change of paradigm:
  taking leave of two tacit assumptions about physical lake forcing} {A change
  of climate provokes a change of paradigm: taking leave of two tacit
  assumptions about physical lake forcing}.{\BBCQ}
\newblock
\APACjournalVolNumPages{International Review of
  Hydrobiology}{93}{4-5}{404--414}.
\PrintBackRefs{\CurrentBib}

\bibitem [\protect \citeauthoryear {%
Lockhoff%
, Zolina%
, Simmer%
\BCBL {}\ \BBA {} Schulz%
}{%
Lockhoff%
\ \protect \BOthers {.}}{%
{\protect \APACyear {2014}}%
}]{%
lockhoff2014evaluation}
\APACinsertmetastar {%
lockhoff2014evaluation}%
\begin{APACrefauthors}%
Lockhoff, M.%
, Zolina, O.%
, Simmer, C.%
\BCBL {}\ \BBA {} Schulz, J.%
\end{APACrefauthors}%
\unskip\
\newblock
\APACrefYearMonthDay{2014}{}{}.
\newblock
{\BBOQ}\APACrefatitle {Evaluation of satellite-retrieved extreme precipitation
  over Europe using gauge observations} {Evaluation of satellite-retrieved
  extreme precipitation over europe using gauge observations}.{\BBCQ}
\newblock
\APACjournalVolNumPages{Journal of climate}{27}{2}{607--623}.
\PrintBackRefs{\CurrentBib}

\bibitem [\protect \citeauthoryear {%
Lovett%
\ \protect \BOthers {.}}{%
Lovett%
\ \protect \BOthers {.}}{%
{\protect \APACyear {2007}}%
}]{%
lovett2007needs}
\APACinsertmetastar {%
lovett2007needs}%
\begin{APACrefauthors}%
Lovett, G\BPBI M.%
, Burns, D\BPBI A.%
, Driscoll, C\BPBI T.%
, Jenkins, J\BPBI C.%
, Mitchell, M\BPBI J.%
, Rustad, L.%
\BDBL {}Haeuber, R.%
\end{APACrefauthors}%
\unskip\
\newblock
\APACrefYearMonthDay{2007}{}{}.
\newblock
{\BBOQ}\APACrefatitle {Who needs environmental monitoring?} {Who needs
  environmental monitoring?}{\BBCQ}
\newblock
\APACjournalVolNumPages{Frontiers in Ecology and the
  Environment}{5}{5}{253--260}.
\PrintBackRefs{\CurrentBib}

\bibitem [\protect \citeauthoryear {%
Ma%
, Cheng%
, Lin%
, Tan%
\BCBL {}\ \BBA {} Zhang%
}{%
Ma%
\ \protect \BOthers {.}}{%
{\protect \APACyear {2019}}%
}]{%
ma2019improving}
\APACinsertmetastar {%
ma2019improving}%
\begin{APACrefauthors}%
Ma, J.%
, Cheng, J\BPBI C.%
, Lin, C.%
, Tan, Y.%
\BCBL {}\ \BBA {} Zhang, J.%
\end{APACrefauthors}%
\unskip\
\newblock
\APACrefYearMonthDay{2019}{}{}.
\newblock
{\BBOQ}\APACrefatitle {Improving air quality prediction accuracy at larger
  temporal resolutions using deep learning and transfer learning techniques}
  {Improving air quality prediction accuracy at larger temporal resolutions
  using deep learning and transfer learning techniques}.{\BBCQ}
\newblock
\APACjournalVolNumPages{Atmospheric Environment}{214}{}{116885}.
\PrintBackRefs{\CurrentBib}

\bibitem [\protect \citeauthoryear {%
Magnuson%
, Benson%
\BCBL {}\ \BBA {} Kratz%
}{%
Magnuson%
\ \protect \BOthers {.}}{%
{\protect \APACyear {1990}}%
}]{%
magnuson1990temporal}
\APACinsertmetastar {%
magnuson1990temporal}%
\begin{APACrefauthors}%
Magnuson, J\BPBI J.%
, Benson, B.%
\BCBL {}\ \BBA {} Kratz, T.%
\end{APACrefauthors}%
\unskip\
\newblock
\APACrefYearMonthDay{1990}{}{}.
\newblock
{\BBOQ}\APACrefatitle {Temporal coherence in the limnology of a suite of lakes
  in Wisconsin, USA} {Temporal coherence in the limnology of a suite of lakes
  in wisconsin, usa}.{\BBCQ}
\newblock
\APACjournalVolNumPages{Freshwater Biology}{23}{1}{145--159}.
\PrintBackRefs{\CurrentBib}

\bibitem [\protect \citeauthoryear {%
Magnuson%
, Crowder%
\BCBL {}\ \BBA {} Medvick%
}{%
Magnuson%
\ \protect \BOthers {.}}{%
{\protect \APACyear {1979}}%
}]{%
magnuson1979temperature}
\APACinsertmetastar {%
magnuson1979temperature}%
\begin{APACrefauthors}%
Magnuson, J\BPBI J.%
, Crowder, L\BPBI B.%
\BCBL {}\ \BBA {} Medvick, P\BPBI A.%
\end{APACrefauthors}%
\unskip\
\newblock
\APACrefYearMonthDay{1979}{}{}.
\newblock
{\BBOQ}\APACrefatitle {Temperature as an ecological resource} {Temperature as
  an ecological resource}.{\BBCQ}
\newblock
\APACjournalVolNumPages{American Zoologist}{19}{1}{331--343}.
\PrintBackRefs{\CurrentBib}

\bibitem [\protect \citeauthoryear {%
\APACciteatitle {Metalearning: Concepts and Systems}}{%
\APACciteatitle {Metalearning: Concepts and Systems}}{%
{\protect \APACyear {2009}}%
}]{%
Brazdil2009}
\APACinsertmetastar {%
Brazdil2009}%
{\BBOQ}\APACrefatitle {Metalearning: Concepts and Systems} {Metalearning:
  Concepts and systems}.{\BBCQ}
\newblock
\APACrefYearMonthDay{2009}{}{}.
\newblock
\BIn{} \APACrefbtitle {Metalearning: Applications to Data Mining}
  {Metalearning: Applications to data mining}\ (\BPGS\ 1--10).
\newblock
\APACaddressPublisher{}{Springer Berlin Heidelberg}.
\newblock
\begin{APACrefDOI} \doi{10.1007/978-3-540-73263-1\_1} \end{APACrefDOI}
\PrintBackRefs{\CurrentBib}

\bibitem [\protect \citeauthoryear {%
Mishra%
, Vu%
, Veettil%
\BCBL {}\ \BBA {} Entekhabi%
}{%
Mishra%
\ \protect \BOthers {.}}{%
{\protect \APACyear {2017}}%
}]{%
mishra2017drought}
\APACinsertmetastar {%
mishra2017drought}%
\begin{APACrefauthors}%
Mishra, A.%
, Vu, T.%
, Veettil, A\BPBI V.%
\BCBL {}\ \BBA {} Entekhabi, D.%
\end{APACrefauthors}%
\unskip\
\newblock
\APACrefYearMonthDay{2017}{}{}.
\newblock
{\BBOQ}\APACrefatitle {Drought monitoring with soil moisture active passive
  (SMAP) measurements} {Drought monitoring with soil moisture active passive
  (smap) measurements}.{\BBCQ}
\newblock
\APACjournalVolNumPages{Journal of Hydrology}{552}{}{620--632}.
\PrintBackRefs{\CurrentBib}

\bibitem [\protect \citeauthoryear {%
Mitchell%
\ \protect \BOthers {.}}{%
Mitchell%
\ \protect \BOthers {.}}{%
{\protect \APACyear {2004}}%
}]{%
mitchell2004multi}
\APACinsertmetastar {%
mitchell2004multi}%
\begin{APACrefauthors}%
Mitchell, K\BPBI E.%
, Lohmann, D.%
, Houser, P\BPBI R.%
, Wood, E\BPBI F.%
, Schaake, J\BPBI C.%
, Robock, A.%
\BDBL {}others%
\end{APACrefauthors}%
\unskip\
\newblock
\APACrefYearMonthDay{2004}{}{}.
\newblock
{\BBOQ}\APACrefatitle {The multi-institution North American Land Data
  Assimilation System (NLDAS): Utilizing multiple GCIP products and partners in
  a continental distributed hydrological modeling system} {The
  multi-institution north american land data assimilation system (nldas):
  Utilizing multiple gcip products and partners in a continental distributed
  hydrological modeling system}.{\BBCQ}
\newblock
\APACjournalVolNumPages{Journal of Geophysical Research:
  Atmospheres}{109}{D7}{}.
\PrintBackRefs{\CurrentBib}

\bibitem [\protect \citeauthoryear {%
Mizukami%
\ \protect \BOthers {.}}{%
Mizukami%
\ \protect \BOthers {.}}{%
{\protect \APACyear {2017}}%
}]{%
mizukami2017towards}
\APACinsertmetastar {%
mizukami2017towards}%
\begin{APACrefauthors}%
Mizukami, N.%
, Clark, M\BPBI P.%
, Newman, A\BPBI J.%
, Wood, A\BPBI W.%
, Gutmann, E\BPBI D.%
, Nijssen, B.%
\BDBL {}Samaniego, L.%
\end{APACrefauthors}%
\unskip\
\newblock
\APACrefYearMonthDay{2017}{}{}.
\newblock
{\BBOQ}\APACrefatitle {Towards seamless large-domain parameter estimation for
  hydrologic models} {Towards seamless large-domain parameter estimation for
  hydrologic models}.{\BBCQ}
\newblock
\APACjournalVolNumPages{Water Resources Research}{53}{9}{8020--8040}.
\PrintBackRefs{\CurrentBib}

\bibitem [\protect \citeauthoryear {%
Pan%
, Yang%
\BCBL {}\ \protect \BOthers {.}}{%
Pan%
\ \protect \BOthers {.}}{%
{\protect \APACyear {2010}}%
}]{%
pan2010survey}
\APACinsertmetastar {%
pan2010survey}%
\begin{APACrefauthors}%
Pan, S\BPBI J.%
, Yang, Q.%
\BCBL {}\ \BOthersPeriod {.}\end{APACrefauthors}%
\unskip\
\newblock
\APACrefYearMonthDay{2010}{}{}.
\newblock
{\BBOQ}\APACrefatitle {A survey on transfer learning} {A survey on transfer
  learning}.{\BBCQ}
\newblock
\APACjournalVolNumPages{TKDE}{}{}{}.
\PrintBackRefs{\CurrentBib}

\bibitem [\protect \citeauthoryear {%
Paniconi%
\ \BBA {} Putti%
}{%
Paniconi%
\ \BBA {} Putti%
}{%
{\protect \APACyear {2015}}%
}]{%
paniconi2015physically}
\APACinsertmetastar {%
paniconi2015physically}%
\begin{APACrefauthors}%
Paniconi, C.%
\BCBT {}\ \BBA {} Putti, M.%
\end{APACrefauthors}%
\unskip\
\newblock
\APACrefYearMonthDay{2015}{}{}.
\newblock
{\BBOQ}\APACrefatitle {Physically based modeling in catchment hydrology at 50:
  Survey and outlook} {Physically based modeling in catchment hydrology at 50:
  Survey and outlook}.{\BBCQ}
\newblock
\APACjournalVolNumPages{Water Resources Research}{51}{9}{7090--7129}.
\PrintBackRefs{\CurrentBib}

\bibitem [\protect \citeauthoryear {%
Patro%
\ \BBA {} Sahu%
}{%
Patro%
\ \BBA {} Sahu%
}{%
{\protect \APACyear {2015}}%
}]{%
patro2015normalization}
\APACinsertmetastar {%
patro2015normalization}%
\begin{APACrefauthors}%
Patro, S.%
\BCBT {}\ \BBA {} Sahu, K\BPBI K.%
\end{APACrefauthors}%
\unskip\
\newblock
\APACrefYearMonthDay{2015}{}{}.
\newblock
{\BBOQ}\APACrefatitle {Normalization: A preprocessing stage} {Normalization: A
  preprocessing stage}.{\BBCQ}
\newblock
\APACjournalVolNumPages{arXiv preprint arXiv:1503.06462}{}{}{}.
\PrintBackRefs{\CurrentBib}

\bibitem [\protect \citeauthoryear {%
Porter%
, Hanson%
\BCBL {}\ \BBA {} Lin%
}{%
Porter%
\ \protect \BOthers {.}}{%
{\protect \APACyear {2012}}%
}]{%
porter2012staying}
\APACinsertmetastar {%
porter2012staying}%
\begin{APACrefauthors}%
Porter, J\BPBI H.%
, Hanson, P\BPBI C.%
\BCBL {}\ \BBA {} Lin, C\BHBI C.%
\end{APACrefauthors}%
\unskip\
\newblock
\APACrefYearMonthDay{2012}{}{}.
\newblock
{\BBOQ}\APACrefatitle {Staying afloat in the sensor data deluge} {Staying
  afloat in the sensor data deluge}.{\BBCQ}
\newblock
\APACjournalVolNumPages{Trends in ecology \& evolution}{27}{2}{121--129}.
\PrintBackRefs{\CurrentBib}

\bibitem [\protect \citeauthoryear {%
{R Core Team}%
}{%
{R Core Team}%
}{%
{\protect \APACyear {2013}}%
}]{%
team2013r}
\APACinsertmetastar {%
team2013r}%
\begin{APACrefauthors}%
{R Core Team}.%
\end{APACrefauthors}%
\unskip\
\newblock
\APACrefYearMonthDay{2013}{}{}.
\newblock
\APACrefbtitle {R: A language and environment for statistical computing.} {R: A
  language and environment for statistical computing.}
\newblock
\APACaddressPublisher{}{Vienna, Austria}.
\PrintBackRefs{\CurrentBib}

\bibitem [\protect \citeauthoryear {%
E\BPBI K.~Read%
\ \protect \BOthers {.}}{%
E\BPBI K.~Read%
\ \protect \BOthers {.}}{%
{\protect \APACyear {2017}}%
}]{%
read2017water}
\APACinsertmetastar {%
read2017water}%
\begin{APACrefauthors}%
Read, E\BPBI K.%
\BCBT {}\ \BOthersPeriod {.}
\end{APACrefauthors}%
\unskip\
\newblock
\APACrefYearMonthDay{2017}{}{}.
\newblock
{\BBOQ}\APACrefatitle {Water quality data for national-scale aquatic research:
  The Water Quality Portal} {Water quality data for national-scale aquatic
  research: The water quality portal}.{\BBCQ}
\newblock
\APACjournalVolNumPages{Water Resources Research}{}{}{}.
\PrintBackRefs{\CurrentBib}

\bibitem [\protect \citeauthoryear {%
J\BPBI S.~Read%
\ \protect \BOthers {.}}{%
J\BPBI S.~Read%
\ \protect \BOthers {.}}{%
{\protect \APACyear {2019}}%
}]{%
read2019process}
\APACinsertmetastar {%
read2019process}%
\begin{APACrefauthors}%
Read, J\BPBI S.%
, Jia, X.%
, Willard, J.%
, Appling, A\BPBI P.%
, Zwart, J\BPBI A.%
, Oliver, S\BPBI K.%
\BDBL {}others%
\end{APACrefauthors}%
\unskip\
\newblock
\APACrefYearMonthDay{2019}{}{}.
\newblock
{\BBOQ}\APACrefatitle {Process-guided deep learning predictions of lake water
  temperature} {Process-guided deep learning predictions of lake water
  temperature}.{\BBCQ}
\newblock
\APACjournalVolNumPages{Water Resources Research}{}{}{}.
\PrintBackRefs{\CurrentBib}

\bibitem [\protect \citeauthoryear {%
J\BPBI S.~Read%
\ \BBA {} Rose%
}{%
J\BPBI S.~Read%
\ \BBA {} Rose%
}{%
{\protect \APACyear {2013}}%
}]{%
read2013physical}
\APACinsertmetastar {%
read2013physical}%
\begin{APACrefauthors}%
Read, J\BPBI S.%
\BCBT {}\ \BBA {} Rose, K\BPBI C.%
\end{APACrefauthors}%
\unskip\
\newblock
\APACrefYearMonthDay{2013}{}{}.
\newblock
{\BBOQ}\APACrefatitle {Physical responses of small temperate lakes to variation
  in dissolved organic carbon concentrations} {Physical responses of small
  temperate lakes to variation in dissolved organic carbon
  concentrations}.{\BBCQ}
\newblock
\APACjournalVolNumPages{Limnology and Oceanography}{58}{3}{921--931}.
\PrintBackRefs{\CurrentBib}

\bibitem [\protect \citeauthoryear {%
J\BPBI S.~Read%
\ \protect \BOthers {.}}{%
J\BPBI S.~Read%
\ \protect \BOthers {.}}{%
{\protect \APACyear {2014}}%
}]{%
read2014simulating}
\APACinsertmetastar {%
read2014simulating}%
\begin{APACrefauthors}%
Read, J\BPBI S.%
, Winslow, L\BPBI A.%
, Hansen, G\BPBI J.%
, Van Den~Hoek, J.%
, Hanson, P\BPBI C.%
, Bruce, L\BPBI C.%
\BCBL {}\ \BBA {} Markfort, C\BPBI D.%
\end{APACrefauthors}%
\unskip\
\newblock
\APACrefYearMonthDay{2014}{}{}.
\newblock
{\BBOQ}\APACrefatitle {Simulating 2368 temperate lakes reveals weak coherence
  in stratification phenology} {Simulating 2368 temperate lakes reveals weak
  coherence in stratification phenology}.{\BBCQ}
\newblock
\APACjournalVolNumPages{Ecological Modelling}{291}{}{142--150}.
\PrintBackRefs{\CurrentBib}

\bibitem [\protect \citeauthoryear {%
Reichstein%
\ \protect \BOthers {.}}{%
Reichstein%
\ \protect \BOthers {.}}{%
{\protect \APACyear {2019}}%
}]{%
reichstein2019deep}
\APACinsertmetastar {%
reichstein2019deep}%
\begin{APACrefauthors}%
Reichstein, M.%
, Camps-Valls, G.%
, Stevens, B.%
, Jung, M.%
, Denzler, J.%
, Carvalhais, N.%
\BCBL {}\ \BOthersPeriod {.}\end{APACrefauthors}%
\unskip\
\newblock
\APACrefYearMonthDay{2019}{}{}.
\newblock
{\BBOQ}\APACrefatitle {Deep learning and process understanding for data-driven
  Earth system science} {Deep learning and process understanding for
  data-driven earth system science}.{\BBCQ}
\newblock
\APACjournalVolNumPages{Nature}{566}{7743}{195}.
\PrintBackRefs{\CurrentBib}

\bibitem [\protect \citeauthoryear {%
Rose%
, Winslow%
, Read%
\BCBL {}\ \BBA {} Hansen%
}{%
Rose%
\ \protect \BOthers {.}}{%
{\protect \APACyear {2016}}%
}]{%
rose2016climate}
\APACinsertmetastar {%
rose2016climate}%
\begin{APACrefauthors}%
Rose, K\BPBI C.%
, Winslow, L\BPBI A.%
, Read, J\BPBI S.%
\BCBL {}\ \BBA {} Hansen, G\BPBI J.%
\end{APACrefauthors}%
\unskip\
\newblock
\APACrefYearMonthDay{2016}{}{}.
\newblock
{\BBOQ}\APACrefatitle {Climate-induced warming of lakes can be either amplified
  or suppressed by trends in water clarity} {Climate-induced warming of lakes
  can be either amplified or suppressed by trends in water clarity}.{\BBCQ}
\newblock
\APACjournalVolNumPages{Limnology and Oceanography Letters}{1}{1}{44--53}.
\PrintBackRefs{\CurrentBib}

\bibitem [\protect \citeauthoryear {%
Roth%
, Nigussie%
\BCBL {}\ \BBA {} Lemann%
}{%
Roth%
\ \protect \BOthers {.}}{%
{\protect \APACyear {2016}}%
}]{%
roth2016model}
\APACinsertmetastar {%
roth2016model}%
\begin{APACrefauthors}%
Roth, V.%
, Nigussie, T\BPBI K.%
\BCBL {}\ \BBA {} Lemann, T.%
\end{APACrefauthors}%
\unskip\
\newblock
\APACrefYearMonthDay{2016}{}{}.
\newblock
{\BBOQ}\APACrefatitle {Model parameter transfer for streamflow and sediment
  loss prediction with SWAT in a tropical watershed} {Model parameter transfer
  for streamflow and sediment loss prediction with swat in a tropical
  watershed}.{\BBCQ}
\newblock
\APACjournalVolNumPages{Environmental Earth Sciences}{75}{19}{1321}.
\PrintBackRefs{\CurrentBib}

\bibitem [\protect \citeauthoryear {%
Samaniego%
, Kumar%
\BCBL {}\ \BBA {} Attinger%
}{%
Samaniego%
\ \protect \BOthers {.}}{%
{\protect \APACyear {2010}}%
}]{%
samaniego2010multiscale}
\APACinsertmetastar {%
samaniego2010multiscale}%
\begin{APACrefauthors}%
Samaniego, L.%
, Kumar, R.%
\BCBL {}\ \BBA {} Attinger, S.%
\end{APACrefauthors}%
\unskip\
\newblock
\APACrefYearMonthDay{2010}{}{}.
\newblock
{\BBOQ}\APACrefatitle {Multiscale parameter regionalization of a grid-based
  hydrologic model at the mesoscale} {Multiscale parameter regionalization of a
  grid-based hydrologic model at the mesoscale}.{\BBCQ}
\newblock
\APACjournalVolNumPages{Water Resources Research}{46}{5}{}.
\PrintBackRefs{\CurrentBib}

\bibitem [\protect \citeauthoryear {%
Schaeffer%
\ \protect \BOthers {.}}{%
Schaeffer%
\ \protect \BOthers {.}}{%
{\protect \APACyear {2018}}%
}]{%
schaeffer2018initial}
\APACinsertmetastar {%
schaeffer2018initial}%
\begin{APACrefauthors}%
Schaeffer, B\BPBI A.%
, Iiames, J.%
, Dwyer, J.%
, Urquhart, E.%
, Salls, W.%
, Rover, J.%
\BCBL {}\ \BBA {} Seegers, B.%
\end{APACrefauthors}%
\unskip\
\newblock
\APACrefYearMonthDay{2018}{}{}.
\newblock
{\BBOQ}\APACrefatitle {An initial validation of Landsat 5 and 7 derived surface
  water temperature for US lakes, reservoirs, and estuaries} {An initial
  validation of landsat 5 and 7 derived surface water temperature for us lakes,
  reservoirs, and estuaries}.{\BBCQ}
\newblock
\APACjournalVolNumPages{International Journal of Remote
  Sensing}{39}{22}{7789--7805}.
\PrintBackRefs{\CurrentBib}

\bibitem [\protect \citeauthoryear {%
Shen%
}{%
Shen%
}{%
{\protect \APACyear {2018}}%
}]{%
shen2018transdisciplinary}
\APACinsertmetastar {%
shen2018transdisciplinary}%
\begin{APACrefauthors}%
Shen, C.%
\end{APACrefauthors}%
\unskip\
\newblock
\APACrefYearMonthDay{2018}{}{}.
\newblock
{\BBOQ}\APACrefatitle {A transdisciplinary review of deep learning research and
  its relevance for water resources scientists} {A transdisciplinary review of
  deep learning research and its relevance for water resources
  scientists}.{\BBCQ}
\newblock
\APACjournalVolNumPages{Water Resources Research}{54}{11}{8558--8593}.
\PrintBackRefs{\CurrentBib}

\bibitem [\protect \citeauthoryear {%
Sivapalan%
\ \protect \BOthers {.}}{%
Sivapalan%
\ \protect \BOthers {.}}{%
{\protect \APACyear {2003}}%
}]{%
sivapalan2003iahs}
\APACinsertmetastar {%
sivapalan2003iahs}%
\begin{APACrefauthors}%
Sivapalan, M.%
, Takeuchi, K.%
, Franks, S.%
, Gupta, V.%
, Karambiri, H.%
, Lakshmi, V.%
\BDBL {}others%
\end{APACrefauthors}%
\unskip\
\newblock
\APACrefYearMonthDay{2003}{}{}.
\newblock
{\BBOQ}\APACrefatitle {IAHS Decade on Predictions in Ungauged Basins (PUB),
  2003--2012: Shaping an exciting future for the hydrological sciences} {Iahs
  decade on predictions in ungauged basins (pub), 2003--2012: Shaping an
  exciting future for the hydrological sciences}.{\BBCQ}
\newblock
\APACjournalVolNumPages{Hydrological sciences journal}{48}{6}{857--880}.
\PrintBackRefs{\CurrentBib}

\bibitem [\protect \citeauthoryear {%
Sola%
\ \BBA {} Sevilla%
}{%
Sola%
\ \BBA {} Sevilla%
}{%
{\protect \APACyear {1997}}%
}]{%
sola1997importance}
\APACinsertmetastar {%
sola1997importance}%
\begin{APACrefauthors}%
Sola, J.%
\BCBT {}\ \BBA {} Sevilla, J.%
\end{APACrefauthors}%
\unskip\
\newblock
\APACrefYearMonthDay{1997}{}{}.
\newblock
{\BBOQ}\APACrefatitle {Importance of input data normalization for the
  application of neural networks to complex industrial problems} {Importance of
  input data normalization for the application of neural networks to complex
  industrial problems}.{\BBCQ}
\newblock
\APACjournalVolNumPages{IEEE Transactions on nuclear
  science}{44}{3}{1464--1468}.
\PrintBackRefs{\CurrentBib}

\bibitem [\protect \citeauthoryear {%
Stanley%
\ \protect \BOthers {.}}{%
Stanley%
\ \protect \BOthers {.}}{%
{\protect \APACyear {2019}}%
}]{%
stanley2019biases}
\APACinsertmetastar {%
stanley2019biases}%
\begin{APACrefauthors}%
Stanley, E\BPBI H.%
, Collins, S\BPBI M.%
, Lottig, N\BPBI R.%
, Oliver, S\BPBI K.%
, Webster, K\BPBI E.%
, Cheruvelil, K\BPBI S.%
\BCBL {}\ \BBA {} Soranno, P\BPBI A.%
\end{APACrefauthors}%
\unskip\
\newblock
\APACrefYearMonthDay{2019}{}{}.
\newblock
{\BBOQ}\APACrefatitle {Biases in lake water quality sampling and implications
  for macroscale research} {Biases in lake water quality sampling and
  implications for macroscale research}.{\BBCQ}
\newblock
\APACjournalVolNumPages{Limnology and Oceanography}{64}{4}{1572--1585}.
\PrintBackRefs{\CurrentBib}

\bibitem [\protect \citeauthoryear {%
Stefan%
, Hondzo%
, Fang%
, Eaton%
\BCBL {}\ \BBA {} McCormick%
}{%
Stefan%
\ \protect \BOthers {.}}{%
{\protect \APACyear {1996}}%
}]{%
stefan1996simulated}
\APACinsertmetastar {%
stefan1996simulated}%
\begin{APACrefauthors}%
Stefan, H.%
, Hondzo, M.%
, Fang, X.%
, Eaton, J.%
\BCBL {}\ \BBA {} McCormick, J.%
\end{APACrefauthors}%
\unskip\
\newblock
\APACrefYearMonthDay{1996}{}{}.
\newblock
{\BBOQ}\APACrefatitle {Simulated long term temperature and dissolved oxygen
  characteristics of lakes in the north-central United States and associated
  fish habitat limits} {Simulated long term temperature and dissolved oxygen
  characteristics of lakes in the north-central united states and associated
  fish habitat limits}.{\BBCQ}
\newblock
\APACjournalVolNumPages{Limnology and Oceanography}{41}{5}{1124--1135}.
\PrintBackRefs{\CurrentBib}

\bibitem [\protect \citeauthoryear {%
Topp%
, Pavelsky%
, Jensen%
, Simard%
\BCBL {}\ \BBA {} Ross%
}{%
Topp%
\ \protect \BOthers {.}}{%
{\protect \APACyear {2020}}%
}]{%
topp2020research}
\APACinsertmetastar {%
topp2020research}%
\begin{APACrefauthors}%
Topp, S\BPBI N.%
, Pavelsky, T\BPBI M.%
, Jensen, D.%
, Simard, M.%
\BCBL {}\ \BBA {} Ross, M\BPBI R.%
\end{APACrefauthors}%
\unskip\
\newblock
\APACrefYearMonthDay{2020}{}{}.
\newblock
{\BBOQ}\APACrefatitle {Research trends in the use of remote sensing for inland
  water quality science: Moving towards multidisciplinary applications}
  {Research trends in the use of remote sensing for inland water quality
  science: Moving towards multidisciplinary applications}.{\BBCQ}
\newblock
\APACjournalVolNumPages{Water}{12}{1}{169}.
\PrintBackRefs{\CurrentBib}

\bibitem [\protect \citeauthoryear {%
Tyralis%
, Papacharalampous%
\BCBL {}\ \BBA {} Langousis%
}{%
Tyralis%
\ \protect \BOthers {.}}{%
{\protect \APACyear {2019}}%
}]{%
tyralis2019brief}
\APACinsertmetastar {%
tyralis2019brief}%
\begin{APACrefauthors}%
Tyralis, H.%
, Papacharalampous, G.%
\BCBL {}\ \BBA {} Langousis, A.%
\end{APACrefauthors}%
\unskip\
\newblock
\APACrefYearMonthDay{2019}{}{}.
\newblock
{\BBOQ}\APACrefatitle {A brief review of random forests for water scientists
  and practitioners and their recent history in water resources} {A brief
  review of random forests for water scientists and practitioners and their
  recent history in water resources}.{\BBCQ}
\newblock
\APACjournalVolNumPages{Water}{11}{5}{910}.
\PrintBackRefs{\CurrentBib}

\bibitem [\protect \citeauthoryear {%
Vanschoren%
}{%
Vanschoren%
}{%
{\protect \APACyear {2018}}%
}]{%
vanschoren2018meta}
\APACinsertmetastar {%
vanschoren2018meta}%
\begin{APACrefauthors}%
Vanschoren, J.%
\end{APACrefauthors}%
\unskip\
\newblock
\APACrefYearMonthDay{2018}{}{}.
\newblock
{\BBOQ}\APACrefatitle {Meta-learning: A survey} {Meta-learning: A
  survey}.{\BBCQ}
\newblock
\APACjournalVolNumPages{arXiv preprint arXiv:1810.03548}{}{}{}.
\PrintBackRefs{\CurrentBib}

\bibitem [\protect \citeauthoryear {%
Vanschoren%
}{%
Vanschoren%
}{%
{\protect \APACyear {2019}}%
}]{%
vanschoren2019meta}
\APACinsertmetastar {%
vanschoren2019meta}%
\begin{APACrefauthors}%
Vanschoren, J.%
\end{APACrefauthors}%
\unskip\
\newblock
\APACrefYearMonthDay{2019}{}{}.
\newblock
{\BBOQ}\APACrefatitle {Meta-learning} {Meta-learning}.{\BBCQ}
\newblock
\BIn{} \APACrefbtitle {Automated Machine Learning} {Automated machine
  learning}\ (\BPGS\ 35--61).
\newblock
\APACaddressPublisher{}{Springer, Cham}.
\PrintBackRefs{\CurrentBib}

\bibitem [\protect \citeauthoryear {%
Wagener%
, Sivapalan%
, Troch%
\BCBL {}\ \BBA {} Woods%
}{%
Wagener%
\ \protect \BOthers {.}}{%
{\protect \APACyear {2007}}%
}]{%
wagener2007catchment}
\APACinsertmetastar {%
wagener2007catchment}%
\begin{APACrefauthors}%
Wagener, T.%
, Sivapalan, M.%
, Troch, P.%
\BCBL {}\ \BBA {} Woods, R.%
\end{APACrefauthors}%
\unskip\
\newblock
\APACrefYearMonthDay{2007}{}{}.
\newblock
{\BBOQ}\APACrefatitle {Catchment classification and hydrologic similarity}
  {Catchment classification and hydrologic similarity}.{\BBCQ}
\newblock
\APACjournalVolNumPages{Geography compass}{1}{4}{901--931}.
\PrintBackRefs{\CurrentBib}

\bibitem [\protect \citeauthoryear {%
Weiss%
, Khoshgoftaar%
\BCBL {}\ \BBA {} Wang%
}{%
Weiss%
\ \protect \BOthers {.}}{%
{\protect \APACyear {2016}}%
}]{%
weiss2016survey}
\APACinsertmetastar {%
weiss2016survey}%
\begin{APACrefauthors}%
Weiss, K.%
, Khoshgoftaar, T\BPBI M.%
\BCBL {}\ \BBA {} Wang, D.%
\end{APACrefauthors}%
\unskip\
\newblock
\APACrefYearMonthDay{2016}{}{}.
\newblock
{\BBOQ}\APACrefatitle {A survey of transfer learning} {A survey of transfer
  learning}.{\BBCQ}
\newblock
\APACjournalVolNumPages{Journal of Big data}{3}{1}{9}.
\PrintBackRefs{\CurrentBib}

\bibitem [\protect \citeauthoryear {%
Wetzel%
\ \BBA {} Likens%
}{%
Wetzel%
\ \BBA {} Likens%
}{%
{\protect \APACyear {2000}}%
}]{%
wetzel2000heat}
\APACinsertmetastar {%
wetzel2000heat}%
\begin{APACrefauthors}%
Wetzel, R\BPBI G.%
\BCBT {}\ \BBA {} Likens, G\BPBI E.%
\end{APACrefauthors}%
\unskip\
\newblock
\APACrefYearMonthDay{2000}{}{}.
\newblock
{\BBOQ}\APACrefatitle {The heat budget of lakes} {The heat budget of
  lakes}.{\BBCQ}
\newblock
\BIn{} \APACrefbtitle {Limnological Analyses} {Limnological analyses}\ (\BPGS\
  45--56).
\newblock
\APACaddressPublisher{}{Springer}.
\PrintBackRefs{\CurrentBib}

\bibitem [\protect \citeauthoryear {%
White%
\ \BBA {} Marshall%
}{%
White%
\ \BBA {} Marshall%
}{%
{\protect \APACyear {2019}}%
}]{%
white2019should}
\APACinsertmetastar {%
white2019should}%
\begin{APACrefauthors}%
White, C\BPBI R.%
\BCBT {}\ \BBA {} Marshall, D\BPBI J.%
\end{APACrefauthors}%
\unskip\
\newblock
\APACrefYearMonthDay{2019}{}{}.
\newblock
{\BBOQ}\APACrefatitle {Should We Care If Models Are Phenomenological or
  Mechanistic?} {Should we care if models are phenomenological or
  mechanistic?}{\BBCQ}
\newblock
\APACjournalVolNumPages{Trends in ecology \& evolution}{34}{4}{276--278}.
\PrintBackRefs{\CurrentBib}

\bibitem [\protect \citeauthoryear {%
Willard%
, Jia%
, Xu%
, Steinbach%
\BCBL {}\ \BBA {} Kumar%
}{%
Willard%
, Jia%
\BCBL {}\ \protect \BOthers {.}}{%
{\protect \APACyear {2020}}%
}]{%
willard2020integrating}
\APACinsertmetastar {%
willard2020integrating}%
\begin{APACrefauthors}%
Willard, J\BPBI D.%
, Jia, X.%
, Xu, S.%
, Steinbach, M.%
\BCBL {}\ \BBA {} Kumar, V.%
\end{APACrefauthors}%
\unskip\
\newblock
\APACrefYearMonthDay{2020}{}{}.
\newblock
{\BBOQ}\APACrefatitle {Integrating physics-based modeling with machine
  learning: A survey} {Integrating physics-based modeling with machine
  learning: A survey}.{\BBCQ}
\newblock
\APACjournalVolNumPages{arXiv preprint arXiv:2003.04919}{}{}{}.
\PrintBackRefs{\CurrentBib}

\bibitem [\protect \citeauthoryear {%
Willard%
, Read%
, Appling%
\BCBL {}\ \BBA {} Oliver%
}{%
Willard%
, Read%
\BCBL {}\ \protect \BOthers {.}}{%
{\protect \APACyear {2020}}%
}]{%
willard2020data}
\APACinsertmetastar {%
willard2020data}%
\begin{APACrefauthors}%
Willard, J\BPBI D.%
, Read, J\BPBI S.%
, Appling, A\BPBI P.%
\BCBL {}\ \BBA {} Oliver, S\BPBI K.%
\end{APACrefauthors}%
\unskip\
\newblock
\APACrefYearMonthDay{2020}{}{}.
\newblock
\APACrefbtitle {Data release: Predicting Water Temperature Dynamics of
  Unmonitored Lakes with Meta Transfer Learning.} {Data release: Predicting
  water temperature dynamics of unmonitored lakes with meta transfer learning.}
\newblock
\APAChowpublished {U.S. Geological Survey - ScienceBase}.
\newblock
\begin{APACrefDOI} \doi{10.5066/P9I00WFR} \end{APACrefDOI}
\PrintBackRefs{\CurrentBib}

\bibitem [\protect \citeauthoryear {%
Winslow%
, Hansen%
, Read%
\BCBL {}\ \BBA {} Notaro%
}{%
Winslow%
\ \protect \BOthers {.}}{%
{\protect \APACyear {2017}}%
}]{%
winslow2017large}
\APACinsertmetastar {%
winslow2017large}%
\begin{APACrefauthors}%
Winslow, L\BPBI A.%
, Hansen, G\BPBI J.%
, Read, J\BPBI S.%
\BCBL {}\ \BBA {} Notaro, M.%
\end{APACrefauthors}%
\unskip\
\newblock
\APACrefYearMonthDay{2017}{}{}.
\newblock
{\BBOQ}\APACrefatitle {Large-scale modeled contemporary and future water
  temperature estimates for 10774 Midwestern US Lakes} {Large-scale modeled
  contemporary and future water temperature estimates for 10774 midwestern us
  lakes}.{\BBCQ}
\newblock
\APACjournalVolNumPages{Scientific data}{4}{}{170053}.
\PrintBackRefs{\CurrentBib}

\bibitem [\protect \citeauthoryear {%
Ying%
, Zhang%
, Huang%
\BCBL {}\ \BBA {} Yang%
}{%
Ying%
\ \protect \BOthers {.}}{%
{\protect \APACyear {2018}}%
}]{%
ying2018transfer}
\APACinsertmetastar {%
ying2018transfer}%
\begin{APACrefauthors}%
Ying, W.%
, Zhang, Y.%
, Huang, J.%
\BCBL {}\ \BBA {} Yang, Q.%
\end{APACrefauthors}%
\unskip\
\newblock
\APACrefYearMonthDay{2018}{}{}.
\newblock
{\BBOQ}\APACrefatitle {Transfer learning via learning to transfer} {Transfer
  learning via learning to transfer}.{\BBCQ}
\newblock
\BIn{} \APACrefbtitle {International Conference on Machine Learning}
  {International conference on machine learning}\ (\BPGS\ 5072--5081).
\PrintBackRefs{\CurrentBib}

\bibitem [\protect \citeauthoryear {%
Zenobi%
\ \BBA {} Cunningham%
}{%
Zenobi%
\ \BBA {} Cunningham%
}{%
{\protect \APACyear {2001}}%
}]{%
zenobi2001using}
\APACinsertmetastar {%
zenobi2001using}%
\begin{APACrefauthors}%
Zenobi, G.%
\BCBT {}\ \BBA {} Cunningham, P.%
\end{APACrefauthors}%
\unskip\
\newblock
\APACrefYearMonthDay{2001}{}{}.
\newblock
{\BBOQ}\APACrefatitle {Using diversity in preparing ensembles of classifiers
  based on different feature subsets to minimize generalization error} {Using
  diversity in preparing ensembles of classifiers based on different feature
  subsets to minimize generalization error}.{\BBCQ}
\newblock
\BIn{} \APACrefbtitle {European Conference on Machine Learning} {European
  conference on machine learning}\ (\BPGS\ 576--587).
\PrintBackRefs{\CurrentBib}

\bibitem [\protect \citeauthoryear {%
Zhong%
, Notaro%
, Vavrus%
\BCBL {}\ \BBA {} Foster%
}{%
Zhong%
\ \protect \BOthers {.}}{%
{\protect \APACyear {2016}}%
}]{%
zhong2016recent}
\APACinsertmetastar {%
zhong2016recent}%
\begin{APACrefauthors}%
Zhong, Y.%
, Notaro, M.%
, Vavrus, S\BPBI J.%
\BCBL {}\ \BBA {} Foster, M\BPBI J.%
\end{APACrefauthors}%
\unskip\
\newblock
\APACrefYearMonthDay{2016}{}{}.
\newblock
{\BBOQ}\APACrefatitle {Recent accelerated warming of the Laurentian Great
  Lakes: Physical drivers} {Recent accelerated warming of the laurentian great
  lakes: Physical drivers}.{\BBCQ}
\newblock
\APACjournalVolNumPages{Limnology and Oceanography}{61}{5}{1762--1786}.
\PrintBackRefs{\CurrentBib}

\end{thebibliography}

\end{document}